\documentclass[sn-apa]{sn-jnl}

\usepackage{bm} 
\usepackage[T1]{fontenc}
\usepackage{tipa} 
\usepackage{textcomp} 
\usepackage{bookmark} 
\usepackage{linguex} 
\usepackage{soul} 

\jyear{2021}
\theoremstyle{thmstyleone}%
\theoremstyle{thmstyletwo}%
\theoremstyle{thmstylethree}%
\begin{document}
\title{Semantic properties of English nominal pluralization}
\subtitle{Insights from word embeddings}
\author*[1]{\fnm{Elnaz} \sur{Shafaei-Bajestan}}\email{elnaz.shafaei-bajestan@uni-tuebingen.de}
\author[1]{\fnm{Masoumeh} \sur{Moradipour-Tari}}\email{masoumeh.moradipour-tari@student.uni-tuebingen.de}
\author[2]{\fnm{Peter} \sur{Uhrig}}\email{peter.uhrig@fau.de}
\author[1]{\fnm{R. Harald} \sur{Baayen}}\email{harald.baayen@uni-tuebingen.de}
\affil[1]{\orgname{Eberhard Karls Universität Tübingen}, \country{Germany}}
\affil[2]{\orgname{Friedrich-Alexander-Universität Erlangen-Nürnberg}, \country{Germany}}

\newpage

\abstract{Semantic differentiation of nominal pluralization is grammaticalized in many languages. For example, plural markers may only be relevant for human nouns. 
English does not appear to make such distinctions. Using distributional semantics, we show that English nominal pluralization exhibits semantic clusters. For instance, pluralization of fruit words is more similar to one another and less similar to pluralization of other semantic classes. Therefore, reduction of the meaning shift in plural formation to the addition of an abstract plural meaning is too simplistic. A semantically informed method, called CosClassAvg, is introduced that outperforms pluralization methods in distributional semantics which assume plural formation amounts to the addition of a fixed plural vector. In comparison with our approach, a method from compositional distributional semantics, called FRACSS, predicted plural vectors that were more similar to the corpus-extracted plural vectors in terms of direction but not vector length. A modeling study reveals that the observed difference between the two predicted semantic spaces by CosClassAvg and FRACSS carries over to how well a computational model of the listener can understand previously unencountered plural forms. 
Mappings from word forms, represented with triphone vectors, to predicted semantic vectors are more productive when CosClassAvg-generated semantic vectors are employed as gold standard vectors instead of FRACSS-generated vectors. 
}


\artnote{This article is under review at \textit{Morphology}.}

\maketitle

\section{Introduction}\label{sec:introduction}


According to \cite{Corbett:2000:Number}, grammatical number in English can be conceptualized as a `feature' \citep[see also][]{Jakobson:1928} with two opposing values: singular and plural.  For English, the formal realization of the nominal plural is the addition of the suffix \textit{-s}, a marker whose absence signals the nominal singular. Textbooks such as \citet{Lieber:2010} and \citet{aronoff2011morphology} take the semantics of plurality for granted and call attention to the way in which the plural is realized. According to \citet[][p. 156]{aronoff2011morphology}, number is generally not marked in the English lexicon, with the exception of pluralia tantum such as {\em pants}.

Two major approaches to morphology, compositional and realizational morphology, analyze English pluralization in subtly different ways. In compositional models, pluralization is argued to be a process that operates on the less complex word (the singular) and results in the more complex word (the plural).  The \textit{morpheme} \textit{-s} is concatenated with the stem, and in parallel, the abstract meaning `plural' that is registered in the \textit{lexical entry} of the \textit{morpheme} \textit{-s}  overrides the singular meaning associated with the stem.

In realizational models, the semantic part of the \textit{word-schema} for plural nouns contains the description `plurality of $x$s' to capture the semantic similarity of all things plural while abstracting away from their differences; hence the use of the variable `$x$'. The schema for plural nouns stands in  correspondence to a schema for singular words with the same semantic symbol `$x$' \citep{Haspelmath:Sims:2010,Booij:2010}.   The operator `plurality of' is assumed to implement the same semantic change for all instances `$x$' to which it is applied.

While compositional models and realizational models are rather different with respect to how they view the relationship between form and meaning at the level of sub-word units,  they are in remarkable agreement when it comes to how the semantics of number for nouns is understood. Both models assume that English plural formation amounts to a change in form that goes hand in hand with the addition of an abstract semantic feature of plurality to a singular word, shifting its denotation from one entity to more than one entity.

As a matter of fact, formal semantics also has assumed a simple theory of plural meaning. \cite{Lasersohn:1995} opens with ``Plurality is a simple notion -- it just means `more than one'.'' Similarly, \cite{Link:1983} in his study of the logical analysis of plurals restricts the domain of entities from which singular nominals take values to atoms (or, ordinary individuals), and plural nominals to non-atomic summations of more than one atom. This interpretation of the plural nominal is often called \textit{exclusive} since it excludes atoms from its denotation \citep{Farkas:deSwart:2010}. It applies to the example \ref{ex:1-a} below. Although an exclusive interpretation of plural meaning is often fine, there are cases where a plural can refer inclusively to one or more entities, as in \ref{ex:1-b}. Here,
\textit{children} in \ref{ex:1-b} is number neutral -- it can mean `one' child or `more than one' children. This is an \textit{inclusive} interpretation that includes atoms and non-atomic sums in its denotation. 

\ex. \a. You're welcome to bring your two children. \label{ex:1-a}
     \b. You're welcome to bring your children. \label{ex:1-b}
     \z. \citep[adapted from][]{Sauerland:Anderssen:Yatsushiro:2005}

Following the maximize presupposition principle in pragmatics \citep{Heim:1991}, \citet{Sauerland:Anderssen:Yatsushiro:2005} and \citet{Liter:Heffner:Schmitt:2017} argue that an exclusive interpretation of a plural form is a consequence of pragmatic inference, which depends on a range of contextual factors \citep{Chemla:2008}. In a situation where the speaker is unsure of the addressee's number of children, \ref{ex:1-b} is appropriate and \ref{ex:2-a} is odd. Assuming that these two sentences are in competition, the use of the singular \textit{child} in \ref{ex:2-a} is blocked because we know from experience that people may have more than one child and the sentence in \ref{ex:1-b} with stronger presuppositions is preferred.

\ex. \textbf{Context}: People can have more than one child.
     \a. *You're welcome to bring your child. \label{ex:2-a}
     \z. \citep[adapted from][]{Sauerland:Anderssen:Yatsushiro:2005} 

In other words, a plural form can be used to denote an unspecified quantity (one, more than one, one or more than one) \citep[see also][]{Mattens:1970} and the exact quantity has to be resolved through interpretation from context. A principal presupposition that underlies this approach to plurality is that the conceptualization of the number feature is orthogonal to the meaning of the nominal phrase. \cite{Sauerland:2003} argues for an additional syntactic (zero) head above the determiner phrase which contains the number feature and its semantic content.
In what follows, we question this presupposition by using distributional semantics (DS).


DS represents words' meanings with high-dimensional numeric vectors, which we will refer to primarily as `semantic vectors' and alternatively as `word embeddings'---as they are known in Natural Language Processing (NLP). Distributional semantics builds on the hypotheses that words that are similar in meaning occur in similar contexts \citep{Rubenstein:Goodenough:1965} and ``words that occur in the same contexts tend to have similar meaning'' \citep{Pantel:2005}.

There are many different ways in which semantic vectors for words can be constructed. Early implementations made use of word by document contingency tables \citep{Landauer:Dumais:1997} or word by context-word contingency tables \citep{Lund:Burgess:1996,Shaoul:Westbury:2010}.  These tables typically yield very high-dimensional vectors with thousands or tens of thousands of dimensions.  By means of dimensionality reduction techniques such as singular value decomposition, the dimensionality of semantic vectors is substantially reduced.  \citet{Landauer:Dumais:1997} recommended 300-dimensional vectors, as in their experience lower-dimensional vectors performed with higher accuracy in a range of tasks such as synonymy detection. 

More recent models make use of artificial neural networks that are trained to predict target words from the words in their immediate context  \citep[e.g., CBOW;][]{Mikolov:Chen:Corrado:Dean:2013} or to predict the words in the immediate context of a target word from that target word \citep[e.g., Skip-gram;][]{Mikolov:Chen:Corrado:Dean:2013}.  A simple three-layer neural network for the Skip-gram model was implemented by \cite{Mikolov:Sutskever:Chen:Corrado:Dean:2013}, using stochastic gradient descent and back-propagation of error.  The model was trained on 100 billion words from the Google News corpus, and the resulting {\tt word2vec} semantic vectors were made available at  \url{https://code.google.com/archive/p/word2vec/}.  

Other word embeddings extend the word2vec methodology by incorporating character n-grams of words \citep[fastText;][]{Bojanowski:2017:fastText} or by  modifying the objective function being optimized \citep[GloVe;][]{Pennington:2014:GloVe}. All these methods extract the semantic vectors purely from textual information. Other studies integrate visual information on top of that and create multi-modal embeddings \citep[e.g.,][]{Shahmohammadi:Lensch:Baayen:2021}. 

Word embeddings are employed advantageously in several tasks within NLP such as named entity recognition, part of speech tagging, sentiment analysis, word sense disambiguation \citep{Wang:Wang:Chen:Wang:Kuo:2019},
and in many areas of psychology and psycholinguistics \citep{Gunther:Rinaldi:Marelli:2019}. \cite{Boleda:2020} discusses their relevance for theoretical linguistics in the areas of diachronic semantic change, polysemy, and the interface between semantics and syntax or semantics and morphology.

The traditional demarcation of morphology and semantics in linguistics is less prominent in DS models. Nevertheless, the statistics used in these models have been shown to encode morphological and syntactic information besides semantic information \citep{Westbury:Hollis:2019}. 
For morphologically related words, measurements from DS models, such as vector similarity, are consistent with human semantic similarity ratings and lexical decision latencies \citep{Rastle:Davis:Marslen-Wilson:Lorraine:2000, Rastle:Davis:New:2004,MoscosodelPradoMartin:Deutsch:Frost:Schreuder:Jong:Baayen:2005, Milin:Kuperman:Kostic:Baayen:2009}. The degree of semantic transparency in English derivation \citep{Marelli:Baroni:2015} and Dutch compounds \citep{Heylen:DeHertog:2012} were explained by DS similarity measures. Findings of \cite{Smolka:Preller:Eulitz:2014} regarding the effect of semantic transparency on morphological priming of German complex verbs were replicated with DS similarity measures by \cite{Pado:Zeller:Snajder:2015} \citep[although][could not fully replicate the latter study]{Shafaei:2017:thesis} \citep[see also][]{Baayen:Smolka:2020}. \cite{Shen:Baayen:2021} find that semantic transparency measured by DS is linked to the productivity of adjective–noun compounds in Mandarin. DS models used in investigating the paradigmatic relation between two Indonesian prefixes \citep{Denistia:Shafaei:Baayen:2021} corroborated the findings of earlier corpus-based analyses. The discriminative lexicon model of \cite{Baayen:Chuang:Shafaei:Blevins:2019} is a computational model of lexical processing, including morphologically complex words, that incorporates insights from distributional semantics for the  representation of word meanings.

In what follows, we utilize word embeddings to study the meaning of English nominal pluralization. DS models from machine learning produce semantic vectors for both singular and plural word forms. However, in order to be useful for the study of morphology, we need to consider additional questions: What does the process of English pluralization, i.e., going from the singular to the plural semantics, mean in this context? How can we model this process? Given a singular meaning, can we conceptualize the plural, and conversely, given the plural meaning, can we conceptualize the singular? As it is more likely that we encounter previously unseen plurals of known singulars, than previously unseen singulars given known plurals, we focus specifically on the productivity of the conceptualization of plural forms and ask: How well can we estimate the semantics of previously unseen plural words? And how does form relate to the estimations for meaning?

In the following section, we first introduce the corpus used in the present study. Sections \ref{sec:realizational-morphology} and \ref{sec:compositional-morphology} investigate the aforementioned questions using methods inspired by  realizational morphology and compositional morphology, respectively. In doing so, we illustrate that the widespread assumption in morphology about plural meaning is too simplistic, and we study alternative approaches  that stay closer to the actual complexity of noun pluralization in English. In section \ref{sec:ldl}, the semantic vectors developed by the best-performing methods (formalizing realizational and compositional morphological theories) are put to use in a word comprehension model \citep[based on the discriminative lexicon model by][]{Baayen:Chuang:Shafaei:Blevins:2019} to study which kind of vectors are optimal for the recognition of previously unseen plural words. A  discussion of the findings concludes the study.

\section{Data}\label{sec:data}

The corpus data used in this study is taken from the NewsScape English Corpus \citep{Uhrig:2018, Uhrig:Habil}. The corpus consists of 269 million tokens from the subtitles of more than 35,000 hours of recordings of US-American TV news collected in the UCLA Library Broadcast NewsScape \citep{Steen:etal:2018}. After capture, the recordings undergo compression, during which the audio channel is recoded into a 96 kbit/sec AAC stream with the Fraunhofer FDK library. For this project, the subtitles collected in the NewsScape text files were processed in an NLP pipeline.

In a first step of this pipeline, sentence splitting was carried out with a purpose-built splitter that takes into account the fact that captions are transmitted in upper case. The resulting sentences were processed with Stanford CoreNLP \citep{Manning:etal:2014:CoreNLP} version 3.7.0, i.e. with PTB3 tokenization. The caseless model included in CoreNLP1 was used to tag every word with a Penn Treebank part-of-speech tag.\footnote{Note that the caseless mode is only available for the left3distsim model but not for the slower but usually better bidirectional tagger model.} Then CoreNLP’s TrueCase annotator was deployed, which overwrites the original text for further processing (preserving the original on a separate level). Dependency Parsing, Named-Entity Recognition and any further processing steps are then based on the case-restored text to ensure consistent results from tools that do not offer caseless models.

After the NLP pipeline, the data was run through a modified version of the forced alignment system Gentle \citep{Ochshorn:Hawkins:2015:Gentle}, which basically runs an automatic speech recognition process with a language model created from the subtitles and then attempts to match the recognized words with the words in the subtitles. The quality of the forced alignment results crucially depends on the accurateness of the transcript it is fed. However, TV subtitles are not exact transcripts. Not only do they often ignore disfluencies such as false starts, but they also omit words and sometimes even change them. The commercials included in the recordings do not systematically come with subtitles either. Thus, Gentle only aligns between 90 and 95 percent of the words in the subtitles successfully, and of these, 92.5\% in a manual evaluation were deemed to be aligned correctly by a human annotator listening to them \citep{Uhrig:Habil}. We have to bear in mind, though that the cutoff points may not have been exact on these words. To increase the quality of the dataset used in the present study, only files where Gentle reported at least 97\% of successfully aligned words were used. 

Words' meanings are represented with semantic vectors of word2vec, which is widely used within NLP and theoretical linguistics. The nearest neighbors of a target word in this semantic space are often semantically similar (e.g., \textit{good} and \textit{great}) or related (\textit{good} and \textit{bad}) words. The top 10 closest neighbors to \textit{Germany} are \textit{German}, \textit{Europe}, \textit{European}, \textit{Sweden}, \textit{Switzerland}, \textit{Austria}, \textit{France}, \textit{Spain}, \textit{Poland}, and \textit{Russia}. \cite{Wang:Wang:Chen:Wang:Kuo:2019} show that similarities computed between pairs of word2vec vectors are highly correlated ($r(2998)=0.72$) with similarity ratings between word pairs obtained from human subjects in the MEN data set \citep{Bruni:Tran:Baroni:2014}, and that word2vec vectors are best performing on syntactic word analogy tasks (see section \ref{sec:realizational-morphology}) juxtaposed with 5 other semantic spaces. \cite{Westbury:Hollis:2019} argue that \cite{Mikolov:Chen:Corrado:Dean:2013}'s approach for training of the word2vec vectors is closely related to the cognitively plausible learning rule of \cite{Rescorla:Wagner:1972}. 

We compiled a \textit{noun pluralization} dataset with 14,699 singular-plural noun pairs from the NewsScape English Corpus with a word2vec vector. Proper names, plurals endings with anything other than an \textit{-s}, plural-singular pairs with the same word-form, and named entities were excluded from the dataset. 

A second set brought together 500,000 tokens of 14,640 orthographic word types from more than 100 television programs in the NewsScape English Corpus with no restriction on their morphological and syntactic categories. The 13,902 words from this set that have a vector available in word2vec compose a \textit{vocabulary} dataset.

\section{Pluralization with realizational morphology} \label{sec:realizational-morphology}

Realizational morphology \citep[also known as word and paradigm morphology;][]{Matthews:1991:Morphology,Stump:2001} posits whole words rather than morphemes as the basic units. A central notion in this theory is the paradigm. In English, the inflectional paradigm for the verb \textit{talk} is \{\textit{talk, talks, talked, talking}\} and for the noun \textit{talk} is \{\textit{talk, talks}\}. Productivity of the lexicon as a system emerges from proportional analogies between words within paradigms, such as $talk:talks :: walk:walks$ (\textit{talk} is to \textit{talks} as \textit{walk} is to \textit{walks}).

\subsection{Proportional analogies with word embeddings} 

Analogical reasoning using word embeddings has been studied for different types of analogical relations including semantic analogies, such as $$man:king::woman:queen,$$ derivational analogies, as in $$quiet:quietly::happy:happily,$$ and inflectional analogies similar to 
\begin{equation}\label{eq:inflectional-analogy-example}
    pen:pens::table:tables.
\end{equation}
\noindent
Various implementations of proportional analogies with word embeddings have been worked out, such as 3CosAdd \citep{Mikolov:Yih:Zweig:2013}, 3CosMul \citep{Levy:Goldberg:2014}, LRCos, and 3CosAvg  \citep{Drozd:Gladkova:Matsuoka:2016}. Performance varies extensively for the different methods and the different types of analogical relations. \cite{Rogers:Drozd:Li:2017} report that, for English, analogical reasoning with embeddings is most successful for inflectional analogies across different methods. These methods are considered below in the context of plural formation.

Most of the aforementioned methods operate on three input vectors to estimate a vector for the target word in a given analogy. For instance, to implement the analogy in (\ref{eq:inflectional-analogy-example}), 3CosAdd predicts a vector for \textit{tables}, labeled $\overrightarrow{tables}_p$, by computing 
\begin{equation}
\overrightarrow{tables}_p= \overrightarrow{pens}-\overrightarrow{pen}+\overrightarrow{table}.
\end{equation}
The word selected as the predicted plural is the word the vector of which is closest to the composed vector, $\overrightarrow{tables}_p$ in (\theequation), in terms of cosine similarity. As a consequence, evaluation of these methods is restricted to predefined analogy test sets such as Google's \citep{Mikolov:Chen:Corrado:Dean:2013} which provide a series of analogies similar to the examples above.
Another limitation of these methods is that their prediction for the target word \textit{tables} highly depends on the prime word pair, here \textit{pen} and \textit{pens}, and not on just the singular word \textit{table} \citep{Rogers:Drozd:Li:2017}.  Thus, the predicted plural vector for \textit{tables} is  different when the prediction builds on another analogy such as $$banana:bananas::table:tables.$$

\noindent
3CosAvg, on the other hand, operates on just one input vector, the vector of the base word. Given the input word \textit{table}, the predicted plural vector by 3CosAvg is 
$$
\overrightarrow{tables}_p=\overrightarrow{table}+\overrightarrow{\textsc{avg shift}}.
$$ 

\noindent
The word selected as the plural form is again, exactly as for 3CosAdd, that word the vector of which is closest to the assembled vector. For plural analogies, \cite{Drozd:Gladkova:Matsuoka:2016} define the average shift vector as
\begin{equation}
    \overrightarrow{\textsc{avg shift}} = \frac{1}{m} \sum_{i=1}^m \Vec{p_i} - \frac{1}{n} \sum_{i=1}^n \Vec{s_i},   
\end{equation}
assuming there are $m$ plural word-forms with vectors $\Vec{p_i}$ and $n$ singular word-forms with vectors $\Vec{s_i}$. The average shift vector is fixed given the data, and represents the semantics of pluralization, just as the plural vector in the discriminative lexicon model \citep{Baayen:Chuang:Shafaei:Blevins:2019} provides a fixed representation for plurality.

\begin{figure}[th]
    \centering
    \includegraphics{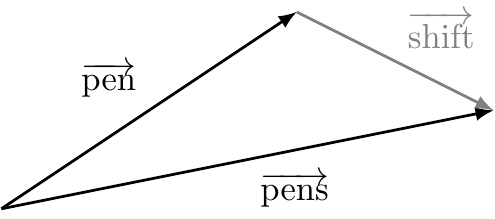}
    \caption{Individual shift vector for the lexeme \textit{pen} is calculated as $\protect\overrightarrow{\text{pens}} - \protect\overrightarrow{\text{pen}}$.}
    \label{fig:shift-vector}
\end{figure}

For a dataset with $m$ plural and $m$ singular word-forms,  the average shift vector, i.e., the difference vector between the average vector of plurals and the average vector of singulars, formulated in (\ref{eq:linear-property-of-offset-vector-line1}), is equal to the average vector of the difference vectors between plurals and singulars, formulated in (\ref{eq:linear-property-of-offset-vector-line3}):
\begin{align}
    \overrightarrow{\textsc{avg shift}} &= \frac{1}{m} \sum_{i=1}^m \Vec{p_i} - \frac{1}{m} \sum_{i=1}^m \Vec{s_i} \label{eq:linear-property-of-offset-vector-line1}\\
    &= \frac{1}{m} (\sum_{i=1}^m \Vec{p_i} -\sum_{i=1}^m \Vec{s_i})\nonumber\\
    &= \frac{1}{m} \sum_{i=1}^m (\Vec{p_i} - \Vec{s_i}). \label{eq:linear-property-of-offset-vector-line3}
\end{align}
\noindent
Henceforth, we refer to the vector $\Vec{p_i} - \Vec{s_i}$ for word $i$ as this word's individual \textit{shift vector}.  Such a shift vector is exemplified in Figure~\ref{fig:shift-vector}.
Importantly, if plural and singular forms for different lexemes are consistently used across similar contexts, as captured by word embeddings, then the difference between individual shift vectors and the average shift vector is expected to be small. 

A range of studies have adopted shift vectors to study the semantics of various lexical relations. For instance, \cite{Roller:Erk:Boleda:Gemma:2014} and \cite{Weeds:2014} used shift vectors for hypernymy detection.  \cite{Bonami:Paperno:2018} used shift vectors to model inflectional and derivational contrasts in French, and \cite{Mickus:Bonami:Paperno:2019} made use of shift vectors for tracing  contrasts in grammatical gender of nouns and adjectives.

\subsection{Individual and average shift vectors} \label{subsec:shift-vectors}

How well does an average shift vector approximate the shifts between individual singulars and their plurals?  To address this question, we investigated what the individual shift vectors look like and whether the average shift vector is representative for the individual shift vectors. For each noun pair in the pluralization dataset represented by word2vec semantic vectors, we first calculated its individual shift vector by subtracting the singular vector from the plural vector.  As a next step, we calculated the length (or magnitude), the direction, and the neighborhood structure of the shift vectors.

\begin{figure}
    \centering
    \includegraphics[width=0.8\textwidth]{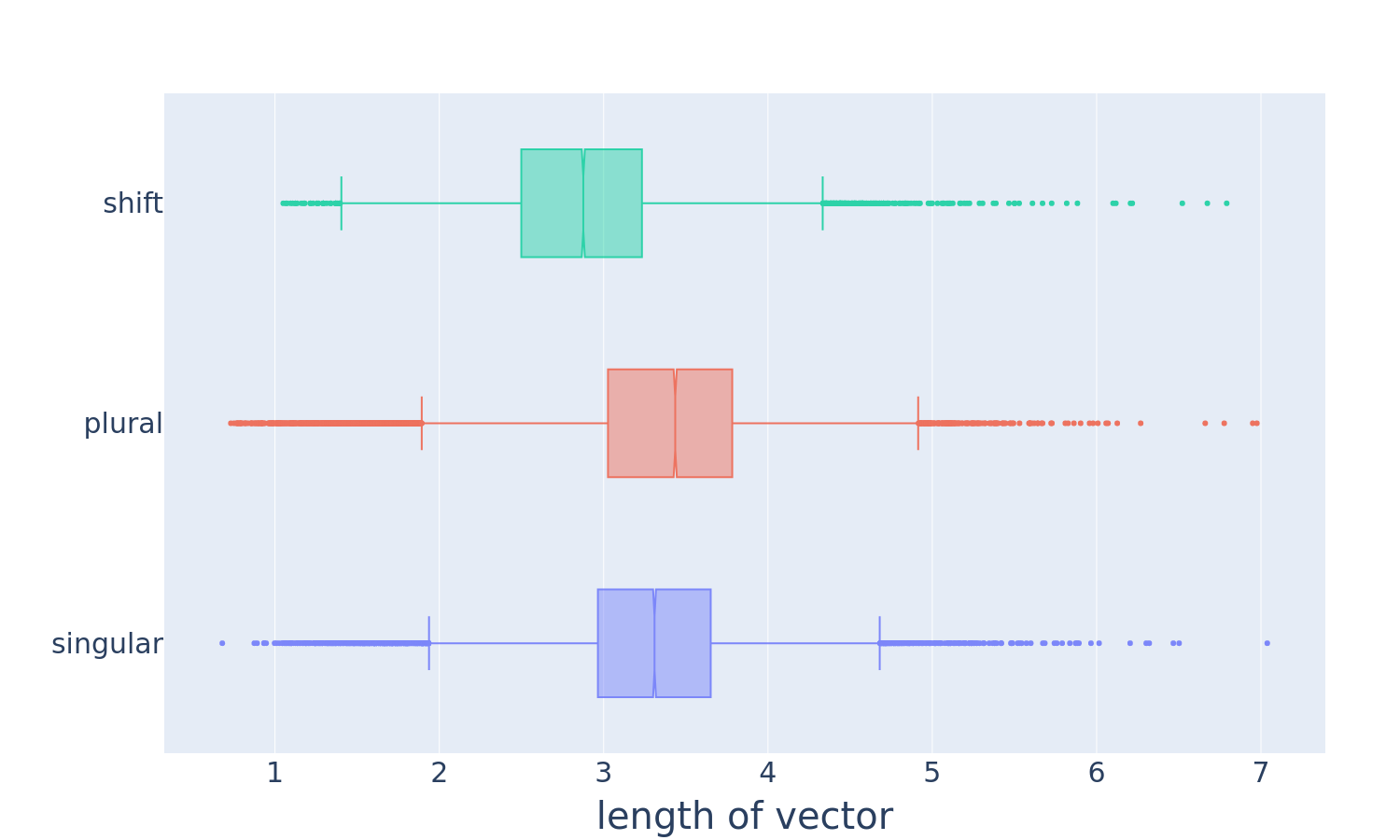}
    \caption{
    Box plots for the length of 14,699 word2vec's singular, plural, and individual shift vectors. 
    }
    \label{fig:boxplot-len-sg-pl-shift}
\end{figure}

We gauged the length of vectors with the $\ell _{2}$ norm, i.e., the Euclidean distance of a vector from the origin. Fig.~\ref{fig:boxplot-len-sg-pl-shift} shows notched box and whiskers plots for the length of singular, plural, and individual shift vectors. Vector lengths differed in the mean  for singular, plural and shift vectors (Friedman test, $\tilde{\chi}^2(2) = 7201$, $p\ll 0.0001$). Pairwise Wilcoxon signed-rank test between groups with Bonferroni correction revealed significant differences in length for all pairwise comparisons (all $p\ll 0.0001$). Plural vectors are, on average, longer than singular vectors (the difference between the medians $\Delta \mathrm{MD}$ is 0.13). This fits well with the intuition that English plurals are semantically more complex than their corresponding singulars. 

Shift vectors are, on average, smaller than the singular ($\Delta \mathrm{MD}=0.43$) and the plural vectors ($\Delta \mathrm{MD}=0.56$), which is only to be expected given that the shift vectors are, by definition, difference vectors.  Although the average length of the shift vectors is smaller than the average lengths of singular or plural vectors, shift vectors turn out to nevertheless be surprisingly long. Their range ($1.1-6.8$,  $\mathrm{MD}=2.88$) is nearly as wide as the ranges of the singular vectors and the plural vectors.  

\begin{figure}
    \centering
    \includegraphics[width=0.8\textwidth]{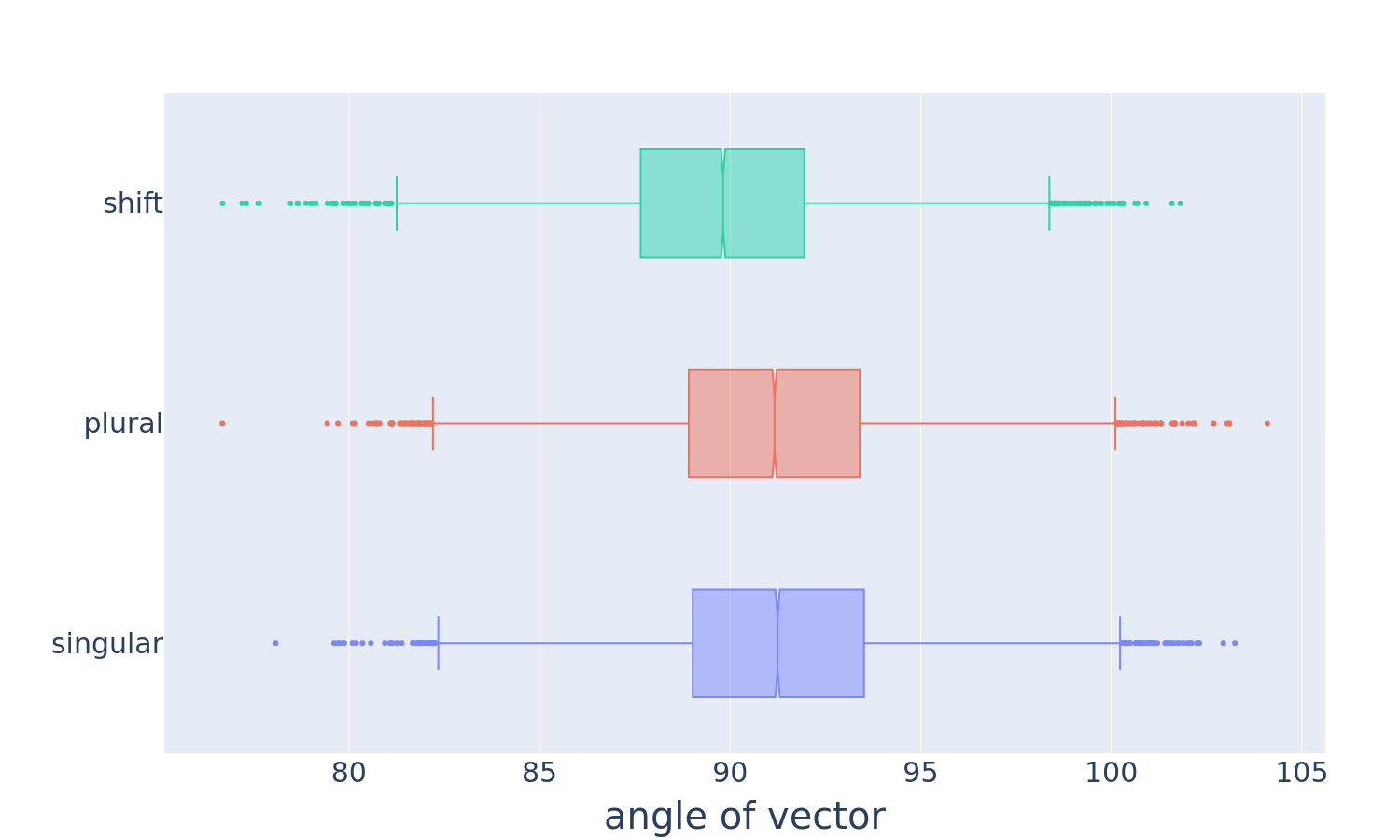}
    \caption{Box plots for the angle of 14,699 word2vec's singular, plural, and individual shift vectors. 
    }
    \label{fig:boxplot-angle-sg-pl-shift}
\end{figure}

We quantified the angles of vectors in word2vec's 300-dimensional vector space with respect to the standard unit vector $\Vec{e}_{300}$ in degrees, using  (\ref{eq:theta}). This 300-dimensional unit vector has a 1 as the last element and zeros elsewhere. Notched boxplots for angle are presented in  Fig.~\ref{fig:boxplot-angle-sg-pl-shift}. The range of angles for shift vectors is even more similar to the ranges of angles of the singular and plural vectors, compared to vector lengths.

\begin{align}\label{eq:theta}
    \theta(\Vec{v}) 
    &= \frac{180}{\pi} (\arccos { \frac{\Vec{v} \cdot \Vec{e}_{300}}{ \left\lVert \Vec{v}\right\rVert_2 \left\lVert \Vec{e}_{300}\right\rVert_2 }) } \nonumber \\
    &= \frac{180}{\pi} (\arccos {\frac {v_{300}}{{\sqrt {\sum \limits _{i=1}^{300}{v_{i}^{2}}}}}})
\end{align}

\begin{figure}
    \centering
    \includegraphics[width=0.85\textwidth]{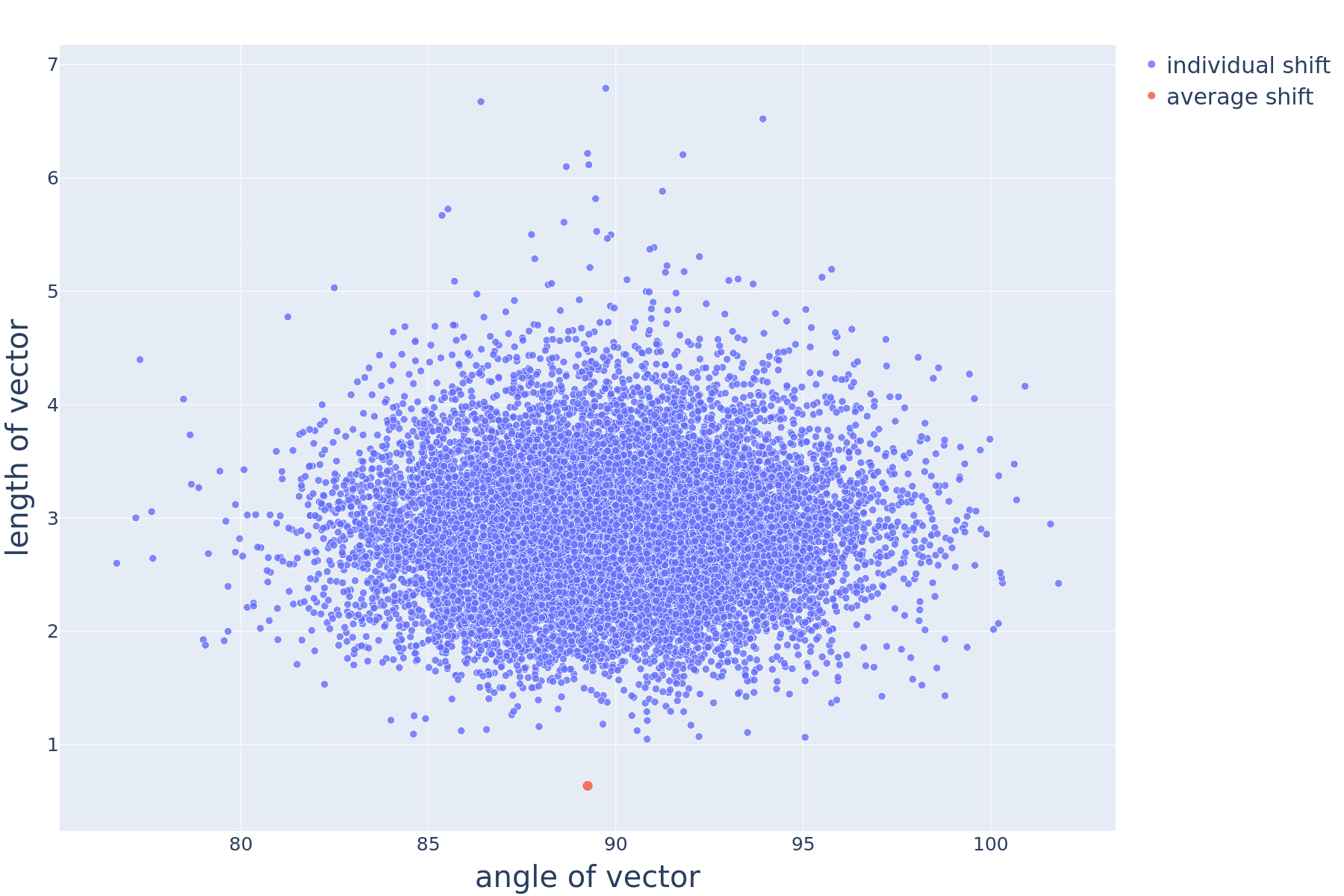}
    \caption{Scatter plot visualizing the relationship between the length, on the y-axis, and angle, on the x-axis, for the individual shift vectors. The isolated red dot below the cloud of all other data points at (89.25, 0.64) belongs to the average shift vector.}
    \label{fig:scatter-r-theta-shift}
\end{figure}

Fig.~\ref{fig:scatter-r-theta-shift} plots the length of shift vectors against their angle. Considerable variability is visible in the length and the angle of individual shift vectors.  The average of a set of vectors radiating from the origin that point in various directions and have various lengths will inevitably end up close to the origin of that vector space.  The average shift vector, in red, at (89.25, 0.64) is smaller than all of the individual shift vectors, and has an $\ell _2$ norm of only 0.64. When such a small vector is added to the singular, it is hardly distinguishable from the singular vector, and at a large distance from the actual corresponding plural vector. 

\begin{figure}
    \centering
    \includegraphics[width=\textwidth]{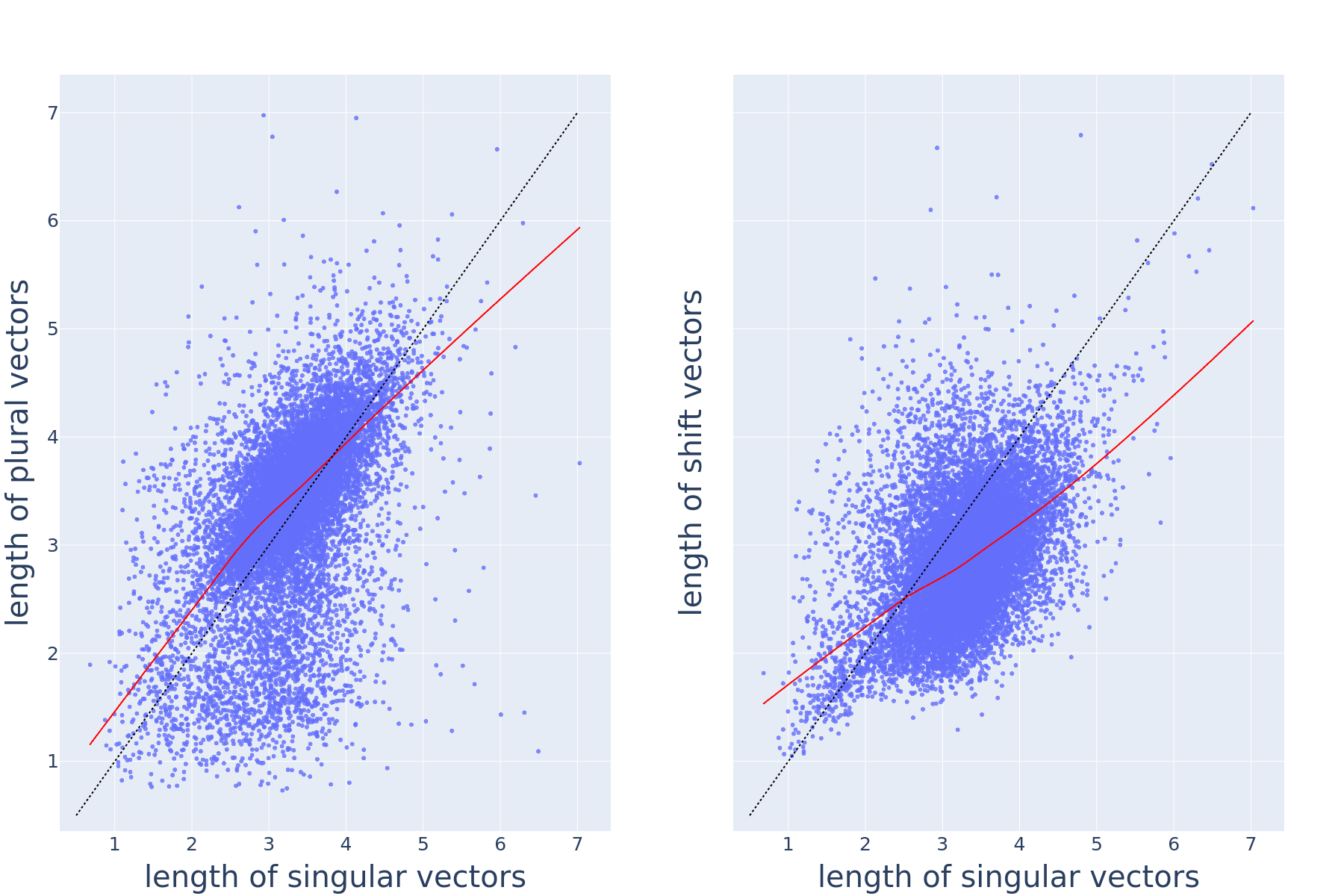}
    \caption{Scatterplots for the length of plural vectors (vertical axis in the left panel) and length of shift vectors (vertical axis in the right panel) against the length of singular vectors (horizontal axis in both plots), with Locally WEighted Scatterplot Smoothing (LOWESS) trend lines in red. The dashed black lines represent the identity line $y=x$.} 
    \label{fig:len-shift-depends-len-sg}
\end{figure}

Upon closer inspection, it turns out that, rather than being random, the set of individual shift vectors exhibits structure. The length of plural vectors increases with the length of their singular vectors, and likewise, the length of shift vectors increases with the length of the singular vectors, as illustrated in Fig.~\ref{fig:len-shift-depends-len-sg}.  From this, we can draw the conclusion that the semantics of shift vectors is changing in close association with the semantics of the singular and plural words.

Given that singular words that have similar semantics have closer vectors, and singular words with less similar meanings have more diverging vectors, we now consider the question of whether the shift vectors themselves show structuring that goes beyond the structure provided at the level of individual lexemes. To address this question, we made use of the t-SNE algorithm for visualizing high-dimensional data \citep{Maaten:Hinton:2008} as implemented in the \texttt{scikit-learn} Python library \citep{scikit-learn}, version 1.0.1, to plot the 300-dimensional shift vectors in a two-dimensional plane.\footnote{Following the recommendations of \cite{vanderMaaten:2021}, we searched the t-SNE's parameter space between possible combinations of \textit{perplexity} (either 10, 15, 20, 25, 30, or 35), \textit{number of iterations} (either 500, 1000, 2000, 3000, or 4000), \textit{random state} (either 1, 12, or 123), and  \textit{initialization} method (either random or PCA) for the t-SNE with the lowest Kullback-Leibler divergence.  The lowest KL-divergence was obtained with the following setting: perplexity $=$ 35, number of iterations = 4000, early exaggeration = 12, random state = 1, learning rate = `auto', metric=`euclidean', and initialization = `random'.}  This visualization technique is known to have a very high chance of recovering the clustering structure present in the input space in the reduced output space \citep{Linderman2019, Arora2018}. 

\begin{figure}
    \centering
    \includegraphics[width=\textwidth]{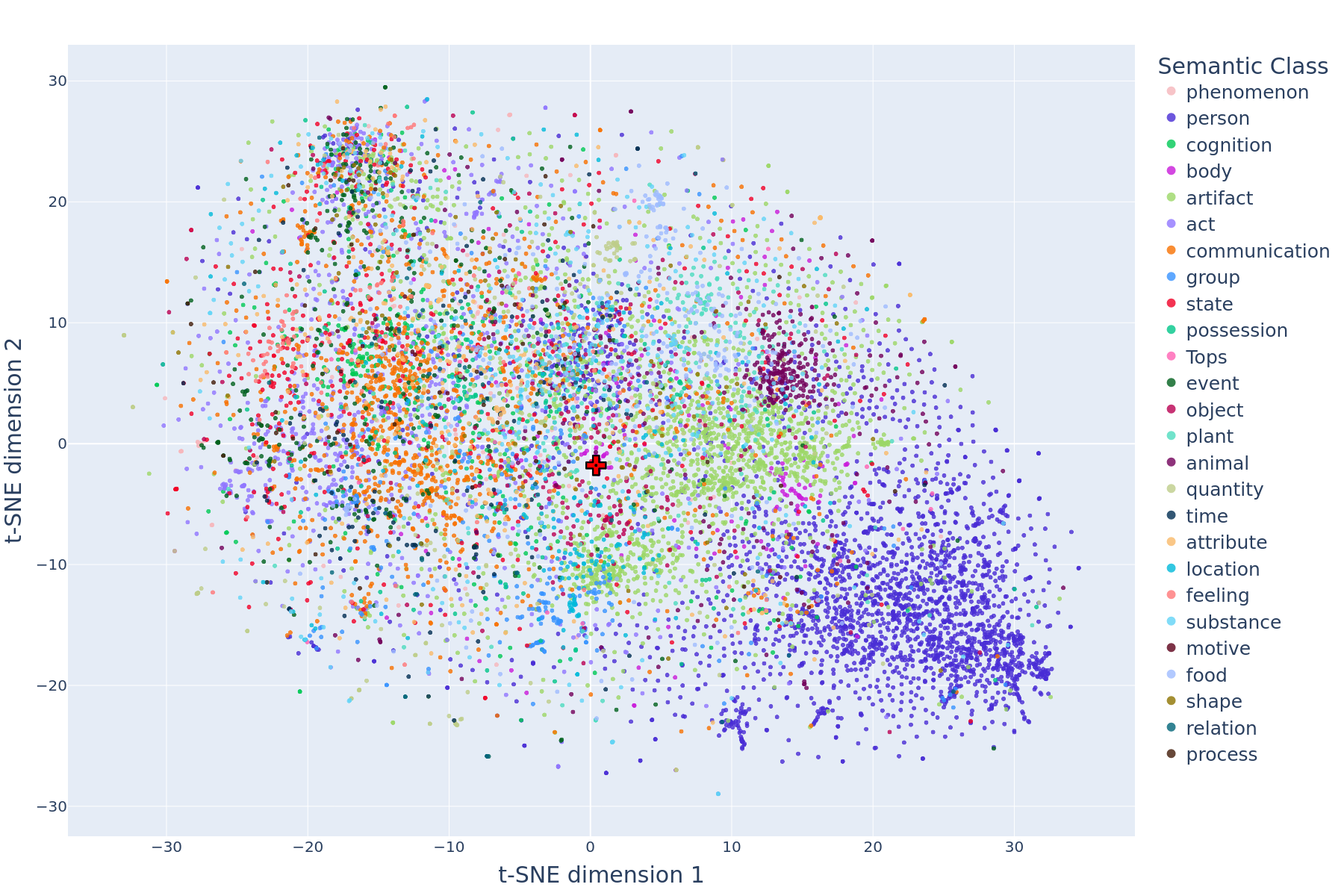}
    \caption{A projection of shift vectors onto a two-dimensional plane using t-SNE reveals semantic clustering. Colors correspond to WordNet supersenses. The average shift vector marked with a red cross, which is located very close to the origin at $(0.4,\, -1.8)$, is blind to this rich structure.  This figure is available as an interactive plot in the supplementary materials.
    }
    \label{fig:tsne-shift-w2v}
\end{figure}

Fig.~\ref{fig:tsne-shift-w2v} presents the scatter of data points in this plane, coloured with the label of the first synset in WordNet \citep{Fellbaum:1998,Miller:1995} for the singular word form. From the 14,699 pairs in our pluralization dataset, 11,749 pairs are found in WordNet and used in the remainder of this study. The labels, indicated in the figure's legend,  often referred to as supersenses,  include 26 broad semantic categories for nouns \citep{Ciaramita:Johnson:2003}. Interestingly, the individual shift vectors form clusters that are reasonably well approximated by the WordNet supersenses. Some supersenses show well-defined clusters, such as \textit{person} towards the bottom right corner of the plane and \textit{animal} towards the top right corner. This indicates that pluralization is similar for nouns denoting animal nouns and is different for nouns denoting persons.  Importantly, the average shift vector (highlighted by a red cross) is located near the origin of this space at $(0.4,\, -1.8)$. 
Interpretation of the t-SNE dimensions is not very straightforward. Preliminary investigation suggests that the first dimension is to a very large extent differentiating between concrete and abstract words (see supplementary materials for details). The second dimension is less interpretable, and rather similar to the first dimension. 

Although some well-defined clusters are present in Fig.~\ref{fig:tsne-shift-w2v}, other clusters show considerable overlap.  This is due to two problems. The first problem is that nouns can have multiple senses. We selected the first sense listed in WordNet, which, according to \citet{Jurafsky:Martin:2021}, is the most frequent sense and hence a strong baseline.  However, inaccuracies are inevitable.   For instance, \textit{strawberry} is assigned the food category while \textit{blueberry} is labeled as a plant.  A related problem is that we have one embedding for all senses, instead of sense-specific embeddings.


The second problem is that the supersenses are often too broad and too over-populated to form semantically coherent groups.  For instance, the supersense \textit{artifact} brings together musical instruments, vehicles, clothes, guns, and buildings among others. In the t-SNE plane, this supersense is found in two distinct regions. The fuzziness of the 26 supersenses is clearly demonstrated by Linear Discriminant Analysis (LDA) given the task of assigning shift vectors to supersenses. From an evaluation of the LDA on all of the data points ($N=11749)$, accuracy and weighted average F-score were both 58.4\% . To put the multiclass classification performance of the LDA into perspective, the weighted average F-score by the LDA is 7 times greater than the weighted average F-score of a baseline classifier that always predicts the most frequent superset. The LDA's performance indicates that on the one hand there is structure and the structure is captured by both a supervised algorithm, i.e., LDA, and an unsupervised algorithm, i.e., t-SNE. On the other hand, it indicates that there is also considerable uncertainty about superset membership.

To address the first problem, one would have to make use of techniques for word-sense disambiguation.  Word sense disambiguation has a very long history in computational linguistics and there are many supervised and unsupervised algorithms designed for this task. One might combine WordNet and FrameNet \citep{Baker:Fillmore:Lowe:1998} annotations as proposed by \cite{Baker:Fellbaum:2009}, train a supervised model \citep[e.g.,][]{Zhong:Ng:2010}, or search for words' nearest neighbors in a contextual word embeddings space \citep{Loureiro:Jorge:2019}.   Given a high-accuracy word sense disambiguation pipeline, one could then apply word sense disambiguation before calculating embeddings using word2vec.  Such a programme, if at all feasible, is outside the scope of the present study.

\begin{figure}
    \centering
    \includegraphics[width=0.9\textwidth]{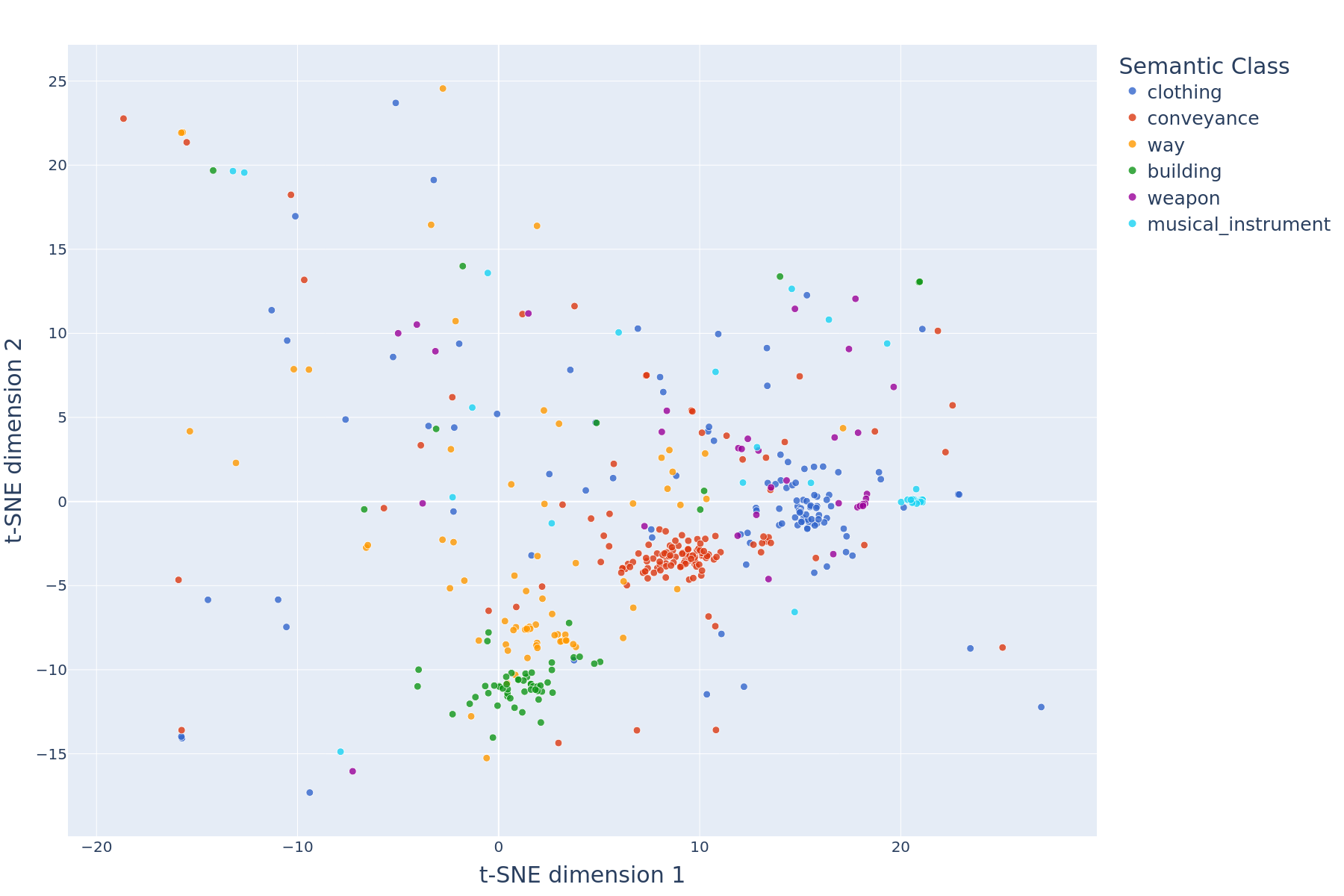}
    \caption{A projection of shift vectors onto a two-dimensional plane using t-SNE for a subset of data points within the supersense \textit{artifact} that form rudimentary sub-clusters such as \textit{clothing} and \textit{musical instruments}.}
    \label{fig:extra}
\end{figure}

The second problem is more straightforward to address.  Instead of using the 26 supersenses shown in Fig.~\ref{fig:tsne-shift-w2v}, we can zoom in on smaller, more semantically homogeneous sense sets.  For instance, the supersense \textit{person} covers 2725 lexemes in our data.  By moving to semantic classes one level below this supersense, we obtain more coherent subsets such as \textit{relative}, \textit{scientist}, and \textit{lover}.  For our pluralization dataset, we constructed a total of 411 classes, by moving zero steps or one step down from the supersenses. On average, a class has 28.6 ($SD=39.8$) members. No class has fewer than 5 members. The most populous class has 481 members. These new semantic classes are more semantically cohesive, as can be seen in Figure~\ref{fig:extra} for a number of sub-classes within the supersense \textit{artifact} as an example. Furthermore, the performance of LDA increased despite the substantial increase in the number of classes. Accuracy and weighted average F-score are both 61\% from an evaluation of an LDA that predicts 411 classes given the shift vectors. In comparison, the weighted average F-score by this model is 189 times greater than the weighted average F-score of a baseline classifier that always predicts the most frequent class.

Although the idea of an abstract semantic representation is appealing, it turns out that a simple average shift vector fails to do justice to the intricate semantic structure that characterizes nominal pluralization in English.  Apparently,  English pluralization is substantially more subtle, and varies systematically with the semantic category (supersense) of a noun.

\subsection{CosClassAvg}\label{subsec:cosclassavg}

This new set of 411 classes, or a similarly cohesive set of classes of semantically highly related words, makes it possible to formalize a new model for plural semantics.  We first calculated the average shift vector for each of the 411 classes. The mean length of these average shift vectors is 1.2, and its standard deviation was 0.3. Compared to the distribution of shift vectors shown in Fig.~\ref{fig:boxplot-len-sg-pl-shift}, both mean and standard deviation are substantially reduced.  The same holds for their angles ($M=89.1$, $\mathrm{SD}=2.6$). This clarifies that by-class shift vectors are more similar to each other  than is the case for the shift vectors in the undifferentiated set of all nouns.

We can now introduce our `CosClassAvg' theory for noun plurals.   Given an input word and its semantic class, the plural vector predicted by CosClassAvg is obtained by taking the singular vector and adding to it the average shift vector for that class. Thus, the vector for \textit{bananas} is predicted using $$\overrightarrow{bananas}_p=\overrightarrow{banana}+\overrightarrow{\textsc{avg-shift}}_{\textsc{fruit}},$$ while the vector for \textit{cars} is predicted based on $$\overrightarrow{cars}_p=\overrightarrow{car}+\overrightarrow{\textsc{avg-shift}}_{\textsc{vehicle}}.$$
We can assess the quality of predicted vectors by inspecting the cosine similarities of a predicted vector with the vectors of all words.  Ideally, the vector that is closest to the predicted vector represents the meaning of the targeted plural.

How well does CosClassAvg perform?  To address this question, we first investigated whether predicted plurals are better differentiated from their singular counterparts.  As our baseline for comparisons, we used the \textsc{Only-b} method introduced in \cite{Linzen:2016}, where \textsc{b} represents the vector for the base word.  This method simply returns the input singular vector, without adding anything to it, as the predicted plural vector. As a consequence, this method will always predict the nearest neighbor in terms of cosine similarity, i.e. the word that is most similar to the base word in the vocabulary.

We calculated the predicted plural vectors for all singular words in our pluralization dataset ($N=11,749$) using 3CosAvg, CosClassAvg, and the baseline method. Many implementations of proportional analogies with word embeddings exclude the input words such as the singular word from the vocabulary as a potential predicted word. However, in an ``honest'' practice, as \cite{Rogers:Drozd:Li:2017} put it, we do not exclude any words from the vocabulary. We therefore compared predicted vectors with a broader set of words covering all 30,497 word-form types in our pluralization and our vocabulary datasets.

\begin{figure}
    \centering
    \includegraphics[width=\textwidth]{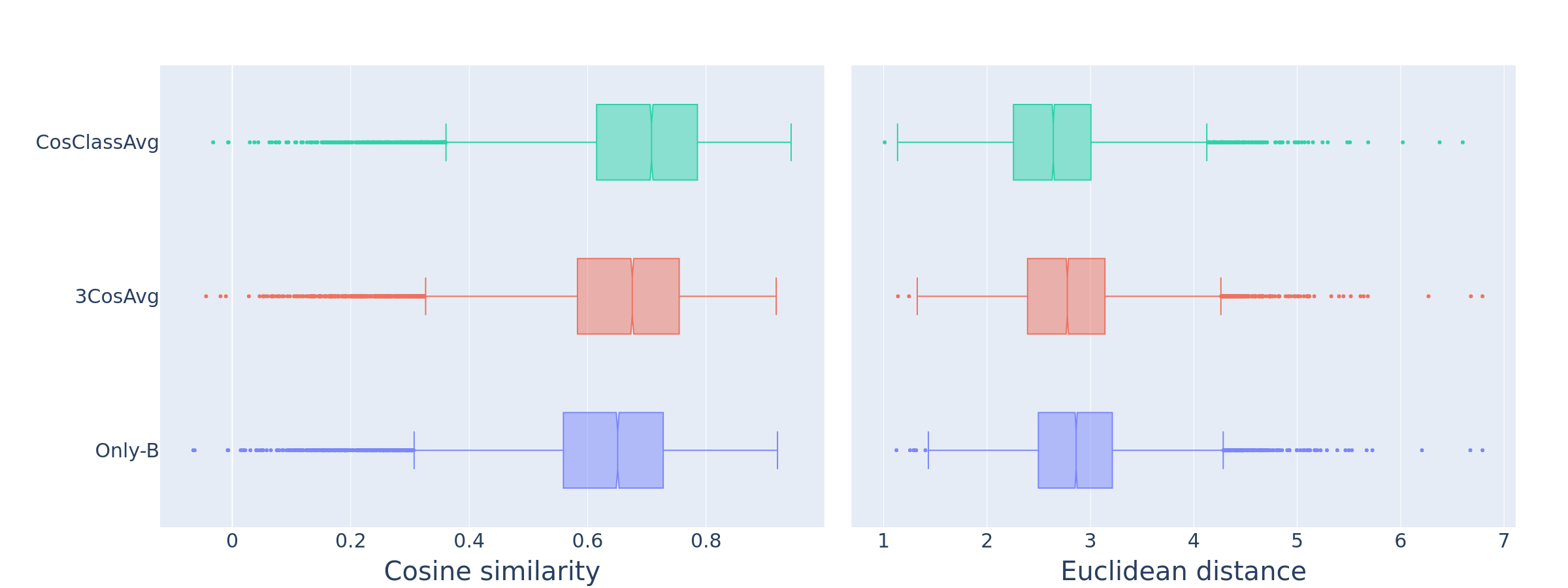}
    \caption{Comparison between the predicted plural vectors and the  corpus-extracted plural vectors using cosine similarity (left panel) and Euclidean distance (right) for the Only-B, the 3CosAvg, and the CosClassAvg method.
    }
    \label{fig:3methods-similarity-hatpl-pl}
\end{figure}

The notched boxplots in Fig.~\ref{fig:3methods-similarity-hatpl-pl} summarize the distributions of cosine similarities (left) and Euclidean distances (right), for the baseline model (Only-B), the 3CosAvg model, and the new CosClassAvg model, of the predicted vectors and the corresponding plural vectors provided by word2vec.  The lowest boxplots in blue produced by the baseline method, indicate that the singular and the plural vectors in word2vec are already astonishingly similar. 
Both 3CosAvg and CosClassAvg improve on the baseline and generate more similar and less distant vectors to the actual plural vector, with CosClassAvg in the lead.

\begin{figure}
    \centering
    \includegraphics[width=\textwidth]{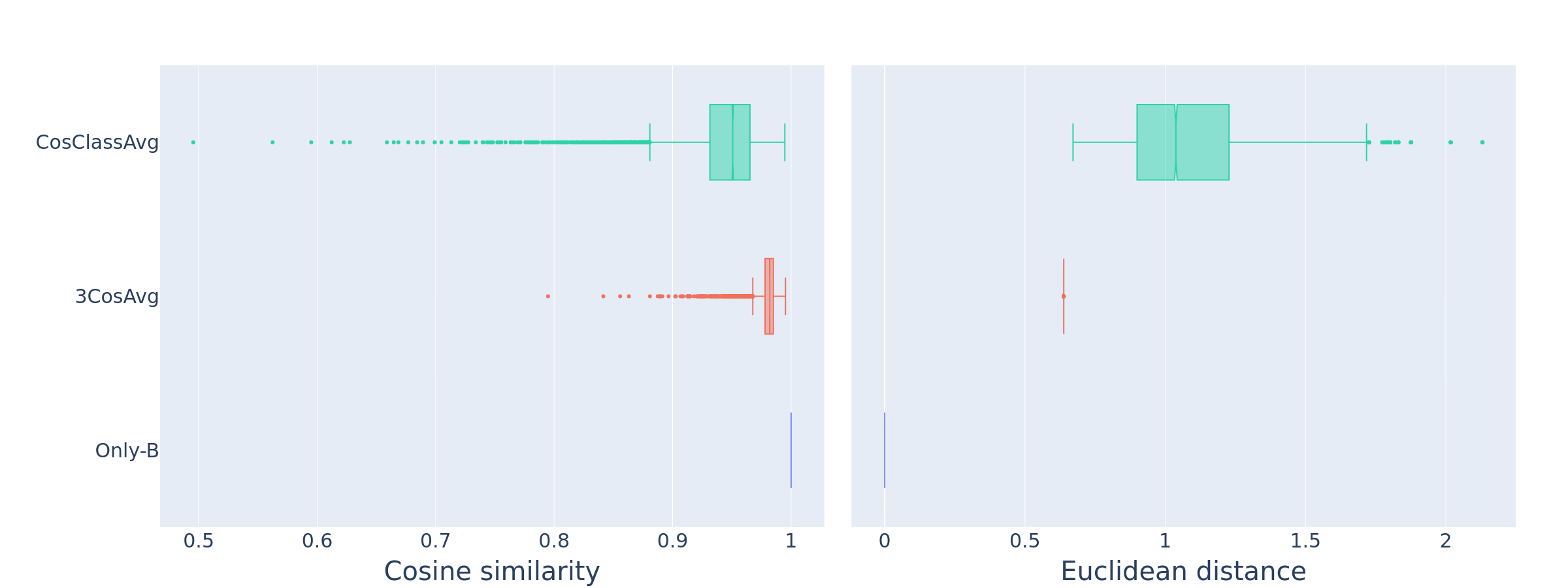}
    \caption{Comparison between the predicted plural vectors and the  singular vectors using cosine similarity (left panel) and Euclidean distance (right) for the Only-B, the 3CosAvg, and the CosClassAvg method.
    }
    \label{fig:3methods-similarity-hatpl-sg}
\end{figure}

For predicted vectors to well approximate the true plural vectors, they should be less close to their corresponding singular vectors. Fig.~\ref{fig:3methods-similarity-hatpl-sg} visualizes cosine similarity to singular vectors and Euclidean distance from singular vectors of the predicted plural vectors. Similarity decreases from one and distance increases from zero with 3CosAvg and CosClassAvg plural vectors. The Euclidean distance between 3CosAvg plural vectors and their singular vectors is always equal to the length of the average shift vector. The length of this overall average shift vector is smaller than the length of any CosClassAvg class-specific shift vector.  

When we use the stringent criterion that any word, including the singular, can be a neighbor of the predicted plural, then performance of both 3CosAvg and CosClassAvg is disappointing. 3CosAvg always selects the singular as closest neighbor, and CosClassAvg only correctly selects 42 plurals (0.4\%).  Although CosClassAvg yields predicted vectors that are further away from their singulars and closer to their plurals, compared to 3CosAvg, predicted plural vectors remain very close to their singular vectors.  Table.~\ref{tab:CosClassAvg-predicts-pl-better} lists the percentages of lexemes for which the targeted plural vector is among the top-n neighbors.  Of the three methods, CosClassAvg clearly outperforms the other two, with percentages ranging from 79\% to 95\%.  In other words, if we relax our criterion and filter out singular vectors as candidates, the accuracy of CosClassAvg is at 79\%.

\begin{table}[ht]
    \begin{center}
    \begin{minipage}{197pt}
    \caption{Percentage of the lexemes ($N=11,749$) for which the plural vector is among the 2 (Top 2), 3 (Top 3), 10 (Top 10), and 20 nearest neighbors (Top 20) of the predicted plural vector.} \label{tab:CosClassAvg-predicts-pl-better}%
        \begin{tabular}{@{}lrrrr@{}}
        \toprule
        Method & Top 2 & Top 3 & Top 10 & Top 20 \\
        \midrule
        Only-B       & 61 & 74 & 88 & 92 \\
        3CosAvg      & 70 & 80 & 91 & 93 \\
        CosClassAvg  & 79 & 86 & 93 & 95 \\
        \botrule
        \end{tabular}
    \end{minipage}
    \end{center}
\end{table}

\subsection{Discussion} 

According to the 3CosAvg method proposed by \citet{Drozd:Gladkova:Matsuoka:2016}, pluralization can be formalized as a function adding an \textit{average shift vector} to the singular vector:
$$
\bm{v}_{\text{pl}} = f_{\text{\tiny{3CosAvg}}}(\bm{v}_{\text{sg}}) = \bm{v}_{\text{sg}} +\bm{v}_{\textsc{pl}}.
$$
We have shown that this formalization of plurality is too simple: shift vectors form semantically motivated clusters.  CosClassAvg brings these classes into a modified function
$$
\bm{v}_{\text{pl}} = g_{\text{\tiny{CosClassAvg}}}(\bm{v}_{\text{sg}}) = \bm{v}_{\text{sg}} +\bm{v}_{\textsc{pl} \mid \text{semantic class}}.
$$
The meaning shift in pluralization is similar for lexemes within a semantic class and is different for lexemes from different semantic classes. CosClassAvg capitalizes on this observation, computes several average shift vectors, one per semantic class, which enables it to generate improved predictions for plurals.

Pluralization with CosClassAvg requires two pieces of information to make a prediction, namely, information on the semantic clusters (and their centroids) and information on the semantic class membership of a given singular noun. The current study shows that, given this information, more precise predictions for plural vectors are obtained. Although outside the scope of this contribution, it may well be possible to develop an end-to-end model that does class induction and pluralization jointly. To that end, semantically cohesive clusters within the shift space may be obtained using unsupervised clustering algorithms. We have shown that clusters found by the unsupervised t-SNE algorithm independently of the WordNet tags, are well-supported by the supervised LDA classification using WordNet tags. In the present approach, we accept the 411 classes as given, leaving it to further research to address the question of how these classes might be grounded in unsupervised learning. 

Regarding the second source of information, we gauged how straightforward it is to classify singular nouns according to their semantics. From a 5-fold stratified cross-validation evaluation of an LDA predicting the 411 semantic classes from \textit{singular} vectors, the mean weighted average F-score was 61\% ($\mathrm{SD} =  0.1$) on the training sets and 32\% ($\mathrm{SD} = 0.5$) on the test sets. The weighted average F-scores by the LDA from the 5 evaluations are on average 190.3 times ($\mathrm{SD} = 0.5$) greater on the training sets and on average 100.1 times ($\mathrm{SD} = 2.3$) greater on the the test sets than the weighted average F-scores of a baseline classifier that always predicts the most frequent class. A straightforward LDA performs quite well under cross-validation. Thus, the classes that we derived from WordNet are to a large extent implicit in the word embeddings.

The CosClassAvg method may also provide enhanced predictivity for human lexical processing, compared to the 3CosAvg method.  For instance, 
\cite{Westbury:Hollis:2019} calculated average vectors for words belonging to different syntactic categories, or containing different derivational affixes, and showed that these average vectors can be leveraged to model human categorization decisions.  
Following their approach, we computed the average vector of plural nouns ($\frac{1}{m} \sum_{i=0}^m \Vec{p_i}$ in equation \ref{eq:linear-property-of-offset-vector-line1}) using the 14,699 plural words in our pluralization dataset introduced in section \ref{sec:data}. Nearest neighbors to the average plural were retrieved among our vocabulary dataset. We replicated \cite{Westbury:Hollis:2019}'s findings for the average plural vector. Within the closest neighbors of the average plural vector, 79\%  are plural nouns. However, other than being plural, these nouns are semantically highly heterogeneous.  If human category decisions are also influenced by the lexical semantics of nouns, more precise predictions can perhaps be obtained by further conditioning on the semantic class of the noun.  We leave this issue to further research. 
  
The clustering of plural shift vectors by semantic class likely reflects differences in how plural objects configure in our (culture-specific) constructions of the world.  Multiple cars occur in different configurations which tend to share alignments, as in parking lots or traffic jams.  Multiple oranges or multiple cherries occur in very different configurations, typically piled up in boxes or on plates, and bananas occur in hands on banana plants and fruit stands.  Apparently, the different properties of the objects that we refer to in the plural are reflected in our language use, as captured by distributional semantics.  However, apples and oranges are more similar than apples and bananas. As a consequence, the vectors predicted by CosClassAvg will always be a bit off for individual words.  This observation necessitates updating the plural semantic function $g$ with an error term, as follows 
$$
\bm{v}_{\text{pl}} = g_{\text{\tiny{CosClassAvg}}}(\bm{v}_{\text{sg}}) = \bm{v}_{\text{sg}} +\bm{v}_{\textsc{pl} \mid \text{semantic class}} + \bm{\epsilon}_\text{lexeme}.
$$
The error vector $\bm{\epsilon}_\text{lexeme}$ represents the lexeme-specific semantics that cannot be captured by the semantic commonalities of the lexeme's semantic class.  In usage-based grammar and corpus linguistics, individual words, including inflected words, have been argued to have their own highly specific usage profiles \citep[see, e.g.,][]{Sinclair:1991}.  `Error' components such as $\bm{\epsilon}_\text{lexeme}$ formalize this important insight.  However, since semantic vectors themselves are measurements, and as such subject to measurement error, we need to add a second error vector representing measurement noise:
$$
\bm{v}_{\text{pl}} = g_{\text{\tiny{CosClassAvg}}}(\bm{v}_{\text{sg}}) = \bm{v}_{\text{sg}} +\bm{v}_{\textsc{pl} \mid \text{semantic class}} + \bm{\epsilon}_\text{lexeme} + \bm{\epsilon}.
$$
Since CosClassAvg decomposes semantic vectors into constituent semantic vectors, it constitutes a `decompositional' or `analytical' method for accounting for inflectional semantics.  In the next section, we compare decompositional CosClassAvg with a compositional method, FRACCS \citep{Marelli:Baroni:2015}.

\section{Pluralization with FRACSS}\label{sec:compositional-morphology}

\cite{Marelli:Baroni:2015}, building on previous research on compositional semantics \citep{Mitchell:Lapata:2008,Baroni:Zamparelli:2010,Lazaridou:Marelli:Zamparelli:Baroni:2013}, proposed to model derivational semantics with the help of a linear transformation that takes the semantic vector of the base word as input, and maps it onto the semantic vector of the corresponding plural using a linear mapping $\bm{B}$: 
$$
\bm{v}_{\text{pl}} = h_{\text{\tiny{FRACSS}}}(\bm{v}_{\text{sg}}) = \bm{v}_{\text{sg}}\bm{B}.
$$
This model, known as the FRACSS model, has been applied to German complex verbs \citep{Gunther:Smolka:Marelli:2019}, and an extended version has been used to study compounding in English and German \citep{Gunther:Marelli:2016, Gunther:Marelli:2019, Marelli:Gagne:Spalding:2017, Gunther:Marelli:Bolte:2020}.   In the following, we apply FRACSS to English plural inflection, and compare its predictions with those of CosClassAvg. 


FRACSS transforms singular vectors into plural vectors using straightforward matrix multiplication.  Let $\bm{X}$ denote a matrix with as row vectors the word embeddings of singulars, and let $\bm{Y}$ denote a matrix with the same number of row and column vectors representing the meanings of the corresponding plurals:
\begin{equation*}
{\displaystyle 
    \bm{X}
    = {\begin{pmatrix}
        x_{1,1} & x_{1,2} & \cdots & x_{1,n}\\
        x_{2,1} & x_{2,2} & \cdots & x_{2,n}\\
        \vdots &\vdots &\ddots & \vdots\\
        x_{t,1}&x_{t,2}&\cdots & x_{t,n}\\
    \end{pmatrix}}, \quad 
    \bm{Y}
    = {\begin{pmatrix}
        y_{1,1} & y_{1,2} & \cdots & y_{1,n}\\
        y_{2,1} & y_{2,2} & \cdots & y_{2,n}\\
        \vdots  & \vdots & \ddots & \vdots\\
        y_{t,1} & y_{t,2} & \cdots & y_{t,n}\\
    \end{pmatrix}}.
}
\end{equation*}

\noindent
The mapping $\bm{B}$ is a $n\times n$ dimensional matrix that satisfies 
$$
\bm{X} \bm{B} = \bm{Y}.
$$
We estimate $\bm{B}$ as follows: 
$$ 
\bm{B} = \bm{X}^+ \bm{Y} = (\bm{X}^T \bm{X})^{-1} \bm{X}^T \bm{Y},
$$
where $\bm{X}^+$ is the pseudo-inverse of $\bm{X}$ and $\bm{X}^T$ is its transpose, and $(.)^{-1}$ denotes a matrix inverse operation. Given $\bm{B}$ and the vector of a singular, the predicted plural vector is given by
\begin{equation*}\label{eq:matrix-multiplication-1vec}
{\displaystyle 
    {\begin{bmatrix}
        x_1 & x_2 & \cdots & x_{n}
    \end{bmatrix}} 
    {\begin{pmatrix}
        b_{1,1} & b_{1,2} & \cdots &b_{1,m}\\
        b_{2,1} & b_{2,2} & \cdots &b_{2,m}\\
        \vdots  &\vdots &\ddots &\vdots\\
        b_{n,1} &b_{n,2}&\cdots &b_{n,m}\\
    \end{pmatrix}} =
    {\begin{bmatrix}
        \hat{y}_1 & \hat{y}_2 & \cdots & \hat{y}_{m}
    \end{bmatrix}}, 
}
\end{equation*}
which, according to the definition of matrix multiplication, implies that
\begin{equation*}
    \hat{y}_j = \sum_{i=1}^{n} b_{i,j} \cdot x_{i}, \qquad {\small \text{for } 1\leq j \leq m }.
\end{equation*}
In other words, the $j$-th element of the semantic vector of a given plural is a weighted sum of the values of its singular vector.

\subsection{Conceptualizing noun plurals with FRACSS} \label{subsec:fracss} 

We estimated the mapping matrix $\bm{B}$ for 90\% of the singular-plural pairs in our pluralization dataset (10,574 pairs) using 300-dimensional word2vec vectors.  The remaining 1,175 word pairs were set aside as held-out testing data.  The resulting 300$\times$300 $\bm{B}$ matrix implements  the change in the meaning of singular words that goes hand in hand with the affixation of the plural \textit{-s}.  With $\bm{B}$ in hand, we can calculate predicted plural vectors for both the training data and the test data.   The model correctly predicts plural forms for 88\% of training items and for 76\% of test items. Clearly, the mapping appears  robust as a memory for seen items and it is also productive for unseen items. 

\begin{figure}
    \centering
    \includegraphics[width=0.9\textwidth]{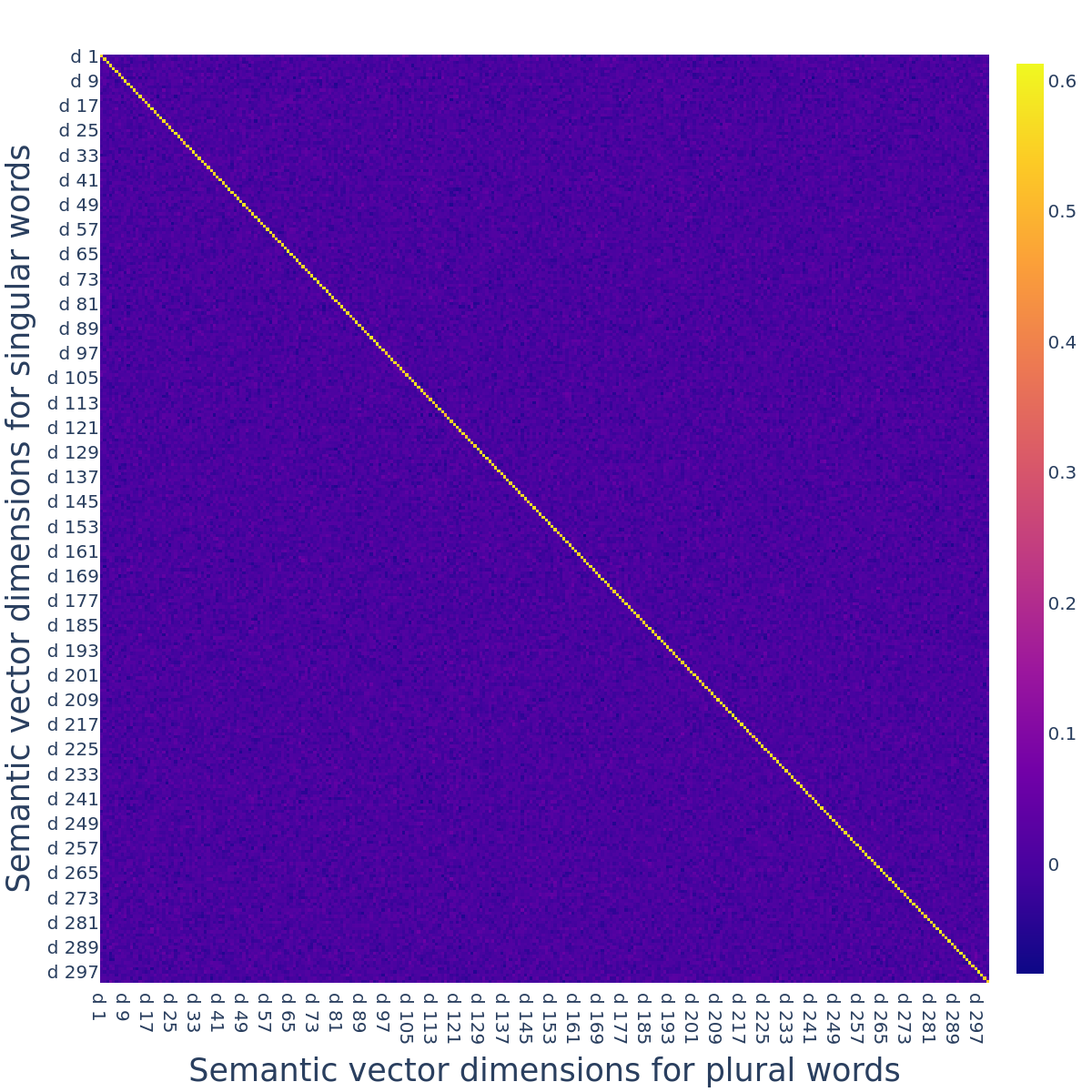}
    \caption{FRACSS matrix for English plural suffix -s.}
    \label{fig:fracss-weights}
\end{figure}

To better understand the performance of the FRACSS model, recall that in section \ref{sec:realizational-morphology} we observed that word2vec's singular and plural vectors are very similar. That is to say, any model for finding a mapping between the singular and the plural space is a-priori in an advantageous position since the relationship between the two spaces is already a given property of the semantic space constructed by word2vec.  In other words, the mapping matrix $\bm{B}$ must be somewhat similar to an identity matrix (i.e., a matrix with ones on the diagonal and zeroes elsewhere).  
The cool-to-warm heat map in Fig.~\ref{fig:fracss-weights}, that visualizes the FRACSS matrix, shows that this is indeed the case. Input vector dimensions are on the vertical axis, indexed by $i$ from 1 to 300, and output vector dimensions, indexed by $j$ from 1 to 300, are on the horizontal axis. The color indicates the magnitude of the value at index $(i,j)$. The value at index $(i,j)$ of this matrix, $b_{i,j}$, shows the association strength between the $i$-th dimension of the singular vectors and the $j$-th dimension of the plural vectors. Association strengths are highest on the diagonal entries of this matrix, which links every singular with its own plural. 

The mean value of the diagonal elements is 0.57 ($SD=0.02$). Barely any structure is evident elsewhere: the mean value of off-diagonal elements is a mere $9.8\times 10^{-5}$ ($SD=0.017$).  We can therefore approximate the effect of multiplication with $\bm{B}$ with a much simpler operation:
\begin{equation}
\hat{\bm{Y}} = 0.57 \bm{X} \bm{I} + \bm{\epsilon},  \bm{\epsilon} \sim {\cal N}_{300}({\bf -0.001}, 0.08\bm{I}),
\end{equation}
\noindent where $\bm{I}$ is the identity matrix, $\bm{\epsilon}$ is a matrix of 300-dimensional random vectors as row vectors all chosen from the same multivariate normal distribution with mean vector $\bf{-0.001}$ (a 300-dimensional vector with -0.001 everywhere) and covariance matrix $0.08\bm{I}$.\footnote{For almost all predicted plural and singular vector pairs, the epsilons were normally distributed with an average mean of -0.001 and an average standard deviation of 0.08 (D'Agostino's $K^2$ departure from normality hypothesis test; $p > 0.001$ for 99.8\% of 11749 tests).} Note that this approximation of $\bm{B}$ predicts that the semantic vectors predicted by FRACSS are shorter in length than their singulars: this follows from the multiplication factor 0.57.

%

How do the FRACSS predicted vectors compare to the vectors predicted by CosClassAvg? To address this question, we first consider similarity evaluated by means of the angle between vectors, and subsequently by means of the Euclidean distance of the corpus-extracted vectors.  The median cosine similarity of predicted and target vectors is 0.75 for FRACSS and 0.71 for CosClassAvg (Wilcoxon signed-rank test $W=65105609.0$, $p \ll 0.0001$ one-tailed, $N=11749$). Furthermore, the median cosine similarity between singular vectors and predicted vectors is 0.87 for FRACSS and 0.95 for CosClassAvg (Wilcoxon signed-rank test $W=649018.0$, $p \ll 0.0001$ one-tailed).

When accuracy is evaluated with the cosine similarity measure, the FRACSS plural vectors are now close enough to the target plural vectors to capture the plural word correctly as the first nearest neighbor in 1520 cases (13\%).  

Similar results are obtained when we use the Euclidean distance measure. The median Euclidean distance to corpus-extracted plural vectors is shorter from predicted vectors for FRACSS at 2.28 in comparison with vectors for CosClassAvg at 2.64 (Wilcoxon signed-rank test $W=1530725.0$, $p \ll 0.0001$ one-tailed). Inversely, the median Euclidean distance between singular vectors and predicted vectors is 1.67 for FRACSS and 1.04 for CosClassAvg (Wilcoxon signed-rank test $W=68556852.0$, $p \ll 0.0001$ one-tailed).

\begin{figure}
    \centering
    \includegraphics[width=\textwidth]{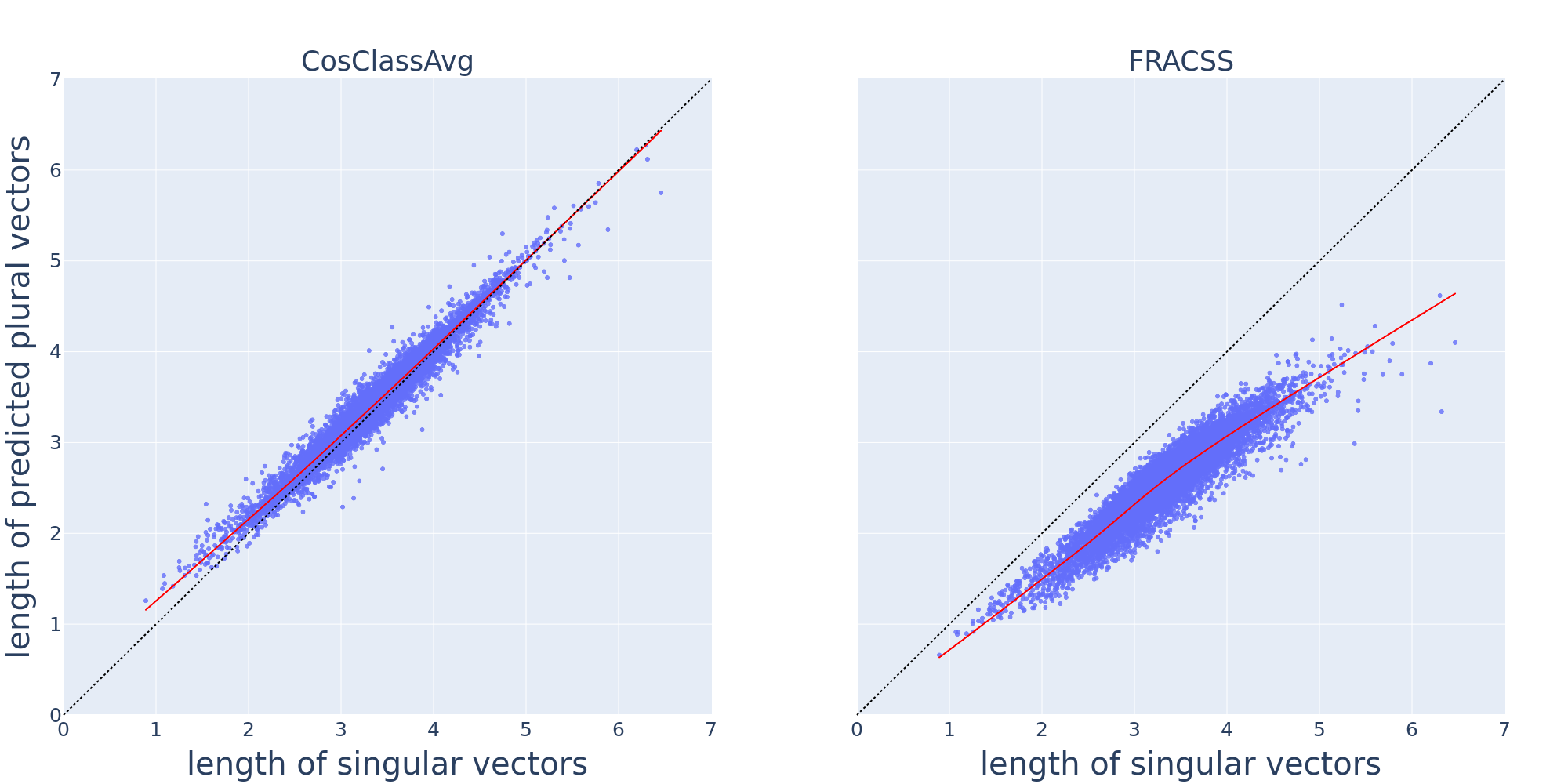}
    \caption{Scatter plots depicting the length of predicted plural vectors, on the vertical axis, versus the length of singular vectors, on the horizontal axis, by CosClassAvg (left panel) and FRACSS (right panel) with the Locally WEighted Scatterplot Smoothing (LOWESS) trend lines in red. The dashed black lines represent the identity line $y=x$.}
    \label{fig:predicted-pl-length}
\end{figure}

Thus far, we have based our evaluation on the angle and distance between vectors. We have seen that FRACSS vectors have smaller angles and shorter distances to plural vectors than CosClassAvg vectors.  What about the Euclidean length of the predicted plural vectors? Fig.~\ref{fig:predicted-pl-length} plots the length of predicted plural vectors against the length of singular vectors, for CosClassAvg (left) and FRACSS (right). For both methods, length of predicted plural vectors increases with the length of singular vectors, similar to the trend observed in Fig.~\ref{fig:len-shift-depends-len-sg} for the length of corpus-extracted plural and singular vectors. However, there is a striking difference. Most plural vectors predicted by CosClassAvg are longer than their singular vector (74\%). By contrast, as anticipated above on the basis of an analysis of the $\bm{B}$ matrix, all plural vectors predicted by FRACSS are shorter than their corresponding singular vectors.\footnote{The signed difference between the length of the target plural vectors and the length of the predicted plural vectors is lower for CosClassAvg compared to FRACSS (Wilcoxon signed-rank test $W=69025375.0, p \ll 0.0001, \mathrm{MD}_{\text{CosClassAvg}}=0.09, \mathrm{MD}_{\text{FRACSS}}=0.89, N=11749$).} However, for the corpus-based actual word2vec vectors,  66\% of the plural vectors are longer than the corresponding singular vectors. 

\subsection{Discussion} 

We have seen that FRACSS outperforms CosClassAvg when evaluation is based on the angle or distance between vectors, but CosClassAvg outperforms FRACSS when we consider vector lengths.  In section~\ref{sec:ldl} we propose another measure for evaluating the relative merits of the two methods.  , we first reflect on some technical and conceptual problems that come with the FRACSS approach.  

One conceptual problem concerns the interpretation of the $\bm{B}$ matrix.  Our t-SNE analysis of shift vectors revealed clustering by semantic class. However, $\bm{B}$ is calculated by evaluating all singulars and plurals simultaneously.  It is an empirical question whether this is advantageous for understanding human lexical processing, an issue we pursue in more detail in section~\ref{sec:ldl}.  If we assume, for the sake of the argument, that FRACSS is a more precise version of CosClassAvg, then CosClassAvg provides us with insight into what FRACSS is actually achieving: semantic-cluster driven local generalization.  In other words, the FRACSS matrix $\bm{B}$ does not represent a single operation of pluralization that is independent and orthogonal to the lexical meaning of the singular. To the contrary, $\bm{B}$ captures a wide range of different local pluralization functions.  

Another conceptual problem concerns the directionality of semantic operations.   \cite{Baroni:Bernardi:Zamparelli:2014} argue that semantic composition is intrinsically asymmetric (compare `water under the bridge' and `bridge under the water'),  and that therefore implementations of semantic functions using matrix multiplication is superior to functions using vector addition.  \cite{Marelli:Baroni:2015} likewise assert that representing affixes as functions over base forms in FRACSS captures this important asymmetry.   However, a FRACSS matrix such as $\bm{B}$ has a pseudo-inverse, and technically it is straightforward to construct a mapping from plurals to singulars. Such a linear mapping trained in the inverse direction on the same data correctly predicts singular semantics for 87\% of training items and 74\% of test items from plural semantics.  We conclude that FRACSS is not intrinsically asymmetric.

There are also some technical issues worth mentioning.  First, updating FRACSS matrices is more costly than updating the CosClassAvg model. Introducing new word pairs, or even a single pair, to the training dataset necessitates re-computation of the whole FRACSS network. For CosClassAvg, a new word pair merely requires re-calibration of the average shift vector for the semantic class of the pertinent lexeme. 

Second, FRACSS requires large numbers of parameters: given vectors of dimension $n$, it requires a mapping with $n^2$ parameters (which can be conceptualized as the beta weights of a multivariate multiple regression model). For the present word2vec vectors, we have no less than 300 $\times$ 300 = 90,000 parameters.  As our current dataset has more than 11,000 datapoints, we have more data than parameters, and FRACSS works just fine. However, when the number of datapoints is substantially less than the number of parameters, the FRACSS approach will overfit the data, and not generalize well. 
For example, in the study of \cite{Lazaridou:Marelli:Zamparelli:Baroni:2013}, 12 out of 18 affixes have fewer than 350 words (training samples), whereas the semantic vectors used had a dimensionality equal to 350.  Likewise, 27 out of 34 affixes studied by \cite{Marelli:Baroni:2015} are trained on fewer observations than their vectors' dimensionality.  

To avoid the problem of overfitting, one could model the complete set of derivational affixes of English with one FRACSS mapping.  Under the assumption that the number of derived words is substantially larger than the dimensionality squared of the embeddings, the model should show good generalization performance.  However, even though data sparsity would no longer be a problem, the model would not be very informative about the semantics of the different affixes. 

In the light of these considerations, we consider how well the two models for conceptualizing plurals, FRACSS and CosClassAvg, perform when integrated into a model of morphology that addresses the mappings between form and meaning, the Discriminative Lexicon (DL) model proposed by \citep{Baayen:Chuang:Shafaei:Blevins:2019}. 

\section{Conceptualization and the mapping from form to meaning} \label{sec:ldl}

Up till now, we have considered how the meanings of singulars and plurals are related to each other, and we have considered two alternative mathematical formalizations of how to conceptualize a plural given a singular.  Both formalizations provide an account of the semantic productivity of pluralization. However, for understanding or producing actual words, we need to consider mappings between form and meaning.  In the ensuing paragraphs, we focus on comprehension, and investigate which of the two formalizations provides semantic vectors that are better aligned with words' forms.  

An initial question is whether words' form representations make systematic contact with the semantic vectors of CosClassAvg and FRACSS. If there exists systematicity between the form space and these semantic spaces, it should be possible to find accurate mappings from forms to meanings, not only for training data, but also for held-out test data.

\subsection{Comprehension with FRACSS and CosClassAvg} 
We model comprehension with the discriminative lexicon model of \cite{Baayen:Chuang:Shafaei:Blevins:2019}. This model makes use of linear mappings from numeric representations of words' forms to numeric representations of words' meanings.  The DL model is well-suited for our purposes as it has been shown to be successful in modeling comprehension of morphologically complex words for various languages \citep{Chuang:Loo:Blevins:Baayen:2020, Heitmeier:Chuang:Baayen:2021, Denistia:Baayen:2022, Heitmeier:Baayen:2020} and, importantly, because it is flexible in terms of which semantic space is selected to represent words' meanings.  Keeping form representations and the representations for singular meanings the same, meaning representations for plurals can be created according to CosClassAvg, or alternatively, according to FRACSS.

For our modeling experiments, we extracted all singular and plural tokens from the vocabulary dataset introduced in section \ref{sec:data}.  This subset comprises 9541 English singular and plural tokens of 8762 unique orthographic word-form types. There are more tokens than types because 728 words have two or more pronunciations in the NewsScape English Corpus. We constructed training data and test data in such a way that plurals in the test data always had the corresponding singular in the training data.  The training data also included plural forms that do not have a corresponding singular in the dataset.  Of all plurals with corresponding singulars, 70\% were assigned to the training data, and 30\% to the testing data.  This resulted in training data comprising 8,507 tokens of 7,886 types, and test data comprising 1034 tokens of 1002 types. Table~\ref{tab:ldl-modeling-data} provides further information on the composition of the training and test sets.  

\begin{table}[ht]
    \begin{center}
    \begin{minipage}{275pt}
    \caption{Number of word-form types and tokens in the datasets used for the DL simulations.}\label{tab:ldl-modeling-data}%
        \begin{tabular}{@{}lrrr@{}} 
        \toprule
        Dataset & Word-form Types & Word-form Tokens \\ 
        \midrule
        \textsc{Training set}\\        
        \hspace{7mm}Singular & 5073 & 5511 \\
        \hspace{7mm}Plural with seen stem & 2253 & 2412 \\
        \hspace{7mm}Plural with unseen stem & 560 & 584 \\
        \textsc{Test set}\\
        \hspace{7mm}Plural with seen stem & 1002 & 1034 \\
        \botrule
        \end{tabular}
    \end{minipage}
    \end{center}
\end{table}

\cite{Heitmeier:Chuang:Baayen:2021} discuss several methods with which numeric representations for word forms can constructed.  In the present study, we make use of numeric form vectors that are based on triphones, i.e., context-sensitive phone units that include information about neighboring segments. For the word \textit{cities}, the triphone cues are \texttt{\#s\textsci}, \texttt{s\textsci t}, \texttt{\textsci ti}, \texttt{tiz}, and \texttt{iz\#}, where the \texttt{\#} symbol is used to denote word boundaries.  For our dataset, there are 6,375 unique triphones.  A word's form vector is defined as a vector with length
6,375 that has values that are either zero or one, depending on whether a triphone is present in a word (1) or not (0).  Words' form vectors can be brought together in a matrix $\bm{C}$ with words on rows and triphones on columns \citep[For form vectors derived from the audio signal, see][]{Shafaei:Tari:Uhrig:Baayen:2021}. As a result, the matrix with word form vectors $\bm{C}$ used for deriving mappings from form to meaning had 8,507 rows and 6,375 columns.

The form vectors for words are based on the phone transcriptions in the NewsScape English Corpus, which are obtained from the Gentle forced aligner. Gentle’s ASR backend is kaldi \citep{Povey:etal:2011:Kaldi}, which is set up to run with a version of the CMUDict machine-readable pronunciation dictionary (\url{https://github.com/cmusphinx/cmudict}), but with information on stress removed.  For various words, the dictionary offers pronunciation variants, such as \texttt{d\_B ae\_I t\_I ah\_E} and \texttt{d\_B ey\_I t\_I ah\_E} for \textit{data}. Here, CMUDict combines ARPABET phone representations with additional information on whether a segment is at the beginning of a word, at an intermediate position, or at the end of a word (\texttt{B}, \texttt{I},  and \texttt{E} respectively). 

We note here that the list of pronunciation variants provided by CMUDict is far from complete. For instance, for \textit{ideology}, it provides the transcription /a\textsci di\textscripta l\textturnv d\textyogh i/ but not the alternative /idi\textscripta l\textturnv d\textyogh i/. Various reduced forms of function words as typically found in spoken language are not represented in the dictionary.  For instance, the conjunction \textit{and} is listed with two variants, /\ae nd/ and /\textturnv nd/, but forms such as /\textturnv n/ or even /n/ are not included.  As a consequence, the representations we used for words' forms may not correspond to the exact way in which these words were actually spoken. 

For evaluating the advantages and disadvantages of semantic vectors based on CosClassAvg and FRACSS, we set up two semantic matrices, $\bm{S}_{\text{\tiny{CosClassAvg}}}$ and $\bm{S}_{\text{\tiny{FRACSS}}}$ that were based on word2vec.  The vectors for singulars were straightforwardly taken from word2vec, but the vectors for plurals were calculated either according to CosClassAvg or according to FRACSS.  The two semantic matrices had 8,507 rows and 300 columns.  We then calculated two 6,375$\times$300 mappings, $\bm{F}_{\text{\tiny{CosClassAvg}}}$ and $\bm{F}_{\text{\tiny{FRACSS}}}$, by solving the equations
\begin{eqnarray*}
\bm{S}_{\text{\tiny{CosClassAvg}}} & = &  \bm{C}\bm{F}_{\text{\tiny{CosClassAvg}}} \\
\bm{S}_{\text{\tiny{FRACSS}}} & = &  \bm{C}\bm{F}_{\text{\tiny{FRACSS}}}.
\end{eqnarray*}
With these the two mappings, we obtained two sets of predicted semantic vectors for the training data:
\begin{eqnarray*}
\hat{\bm{S}}_{\text{\tiny{CosClassAvg}}} & = &  \bm{C}\bm{F}_{\text{\tiny{CosClassAvg}}} \\
\hat{\bm{S}}_{\text{\tiny{FRACSS}}} & = &  \bm{C}\bm{F}_{\text{\tiny{FRACSS}}}.
\end{eqnarray*}
Given the form vectors of the held-out plurals, which we collect as the row vectors of a form matrix $\bm{C}_{\text{\tiny{test}}}$, we also obtain two matrices with predicted plurals:
\begin{eqnarray*}
\hat{\bm{S}}_{\text{\tiny{CosClassAvg}}, \text{\tiny{test}}} & = &  \bm{C}{\text{\tiny{test}}}\bm{F}_{\text{\tiny{CosClassAvg}}} \\
\hat{\bm{S}}_{\text{\tiny{FRACSS}},\text{\tiny{test}} } & = &  \bm{C}{\text{\tiny{test}}}\bm{F}_{\text{\tiny{FRACSS}}}.
\end{eqnarray*}
Prediction accuracy was evaluated by inspecting which gold-standard row vector is closest to the corresponding predicted semantic vector in terms of Pearson's correlation coefficient. If these vectors belong to the same word (i.e., they have the same row index), prediction is taken to be accurate.  In the same way, we can check whether the gold-standard vector is among the top $n$ nearest semantic neighbors. Making a prediction for a given test token always involves choosing among 7,886 $+$ 1 different semantic vectors---the semantic vectors for the word types in the training set plus the semantic vector for the current test word.
Henceforth, we will refer to the DL model with FRACSS vectors as DL-FRACSS and the model with CosClassAvg embeddings as DL-CosClassAvg.

\begin{figure}
    \centering
    \includegraphics[width=0.95\textwidth]{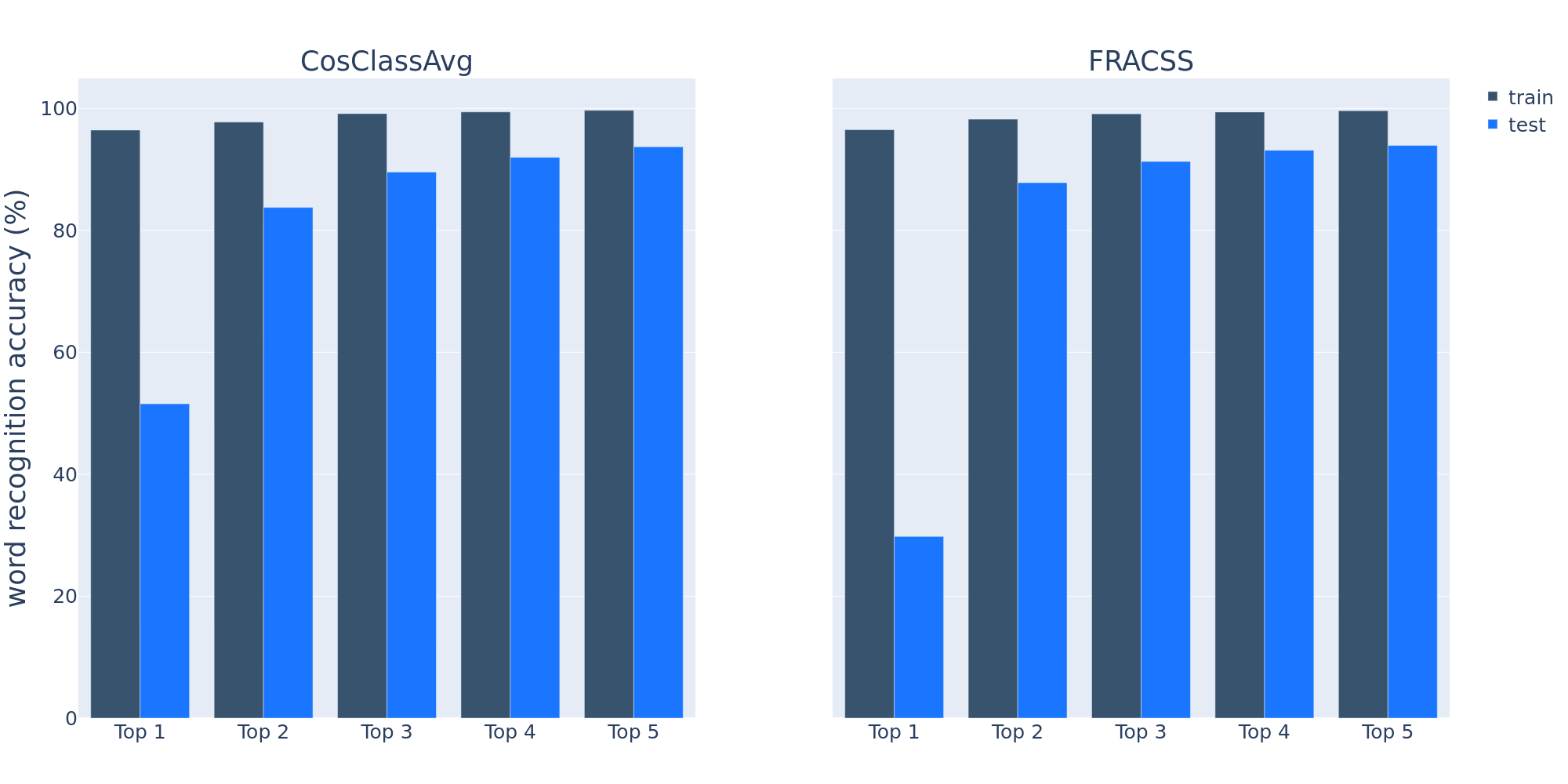}
    \caption{Accuracy of word recognition (\%) on the training set ($N=8507$) and on the test set ($N=1034$) by DL-CosClassAvg on the left panel and by DL-FRACSS on the right panel.}
    \label{fig:ldl-top5-accuracy}
\end{figure}

Fig.~\ref{fig:ldl-top5-accuracy} presents the top 1 to top 5 accuracy of word recognition evaluated on the training set in dark bars and on the test set in light bars.  Recognition accuracy on the training set by both models is 96\% for models' top 1 predictions and increases to almost 100\% as we consider top 2 to top 5 predicted words.  With respect to the test data, DL-CosClassAvg outperforms DL-FRACCS by a wide margin in terms of accuracy (top 1)\footnote{The median correlation between the predicted semantic vectors and the target semantic vector is larger for the DL-CosClassAvg model compared to DL-FRACSS ($W=214575.0$, $p < 0.0001$, $\mathrm{MD}_{\text{DL-CosClassAvg}}=0.78$,  $\mathrm{MD}_{\text{DL-FRACSS}}=0.77$, $N=1034$). }, whereas DL-FRACSS has slightly better performance when the top 2 or top 3 candidates are considered.

Recall that our dataset contains words with multiple pronunciations.  The random selection for inclusion in the held-out dataset of seen-stem plural words may result in either having no instances of the plural word in the training set (e.g., both pronunciations recorded for \textit{reports} occur in the test set), or having one pronunciation in the training data and another pronunciation in the test set (e.g., \textit{results} is trained on /\textturnr \textsci z\textturnv lts/ and tested on /\textturnr iz\textturnv lts/). DL-CosClassAvg recognizes at least one instance of a word in the test set correctly for 63\% of words with multiple pronunciations ($N=155$). DL-FRACSS performs slightly worse at 46\%.

\begin{figure}
    \centering
    \includegraphics[width=0.7\textwidth]{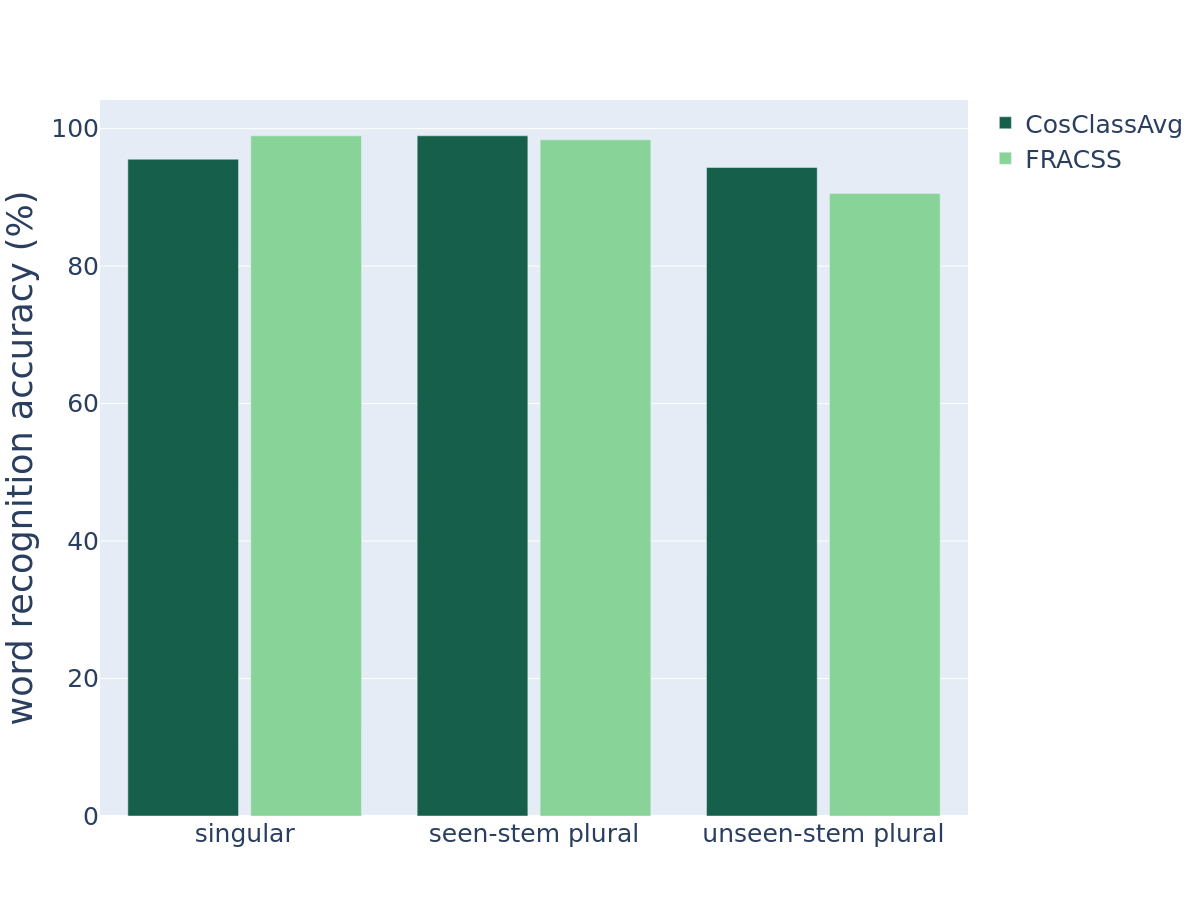}
    \caption{Recognition accuracy on the training set for the singular words, the plural words with a corresponding singular in the training set (seen-stem), and the plural words without a singular in the training set (unseen-stem) by DL-CosClassAvg shown in dark green bars and by DL-FRACSS in light green bars.}
    \label{fig:ldl-top1-accuracy-train-categories}
\end{figure}

Fig.~\ref{fig:ldl-top1-accuracy-train-categories} summarizes model accuracy for the training data.  DL-FRACSS is slightly better at recognizing singulars, whereas DL-CosClassAvg performs slightly better for plurals with unseen stems.

\begin{figure}
    \centering
    \includegraphics[width=0.7\textwidth]{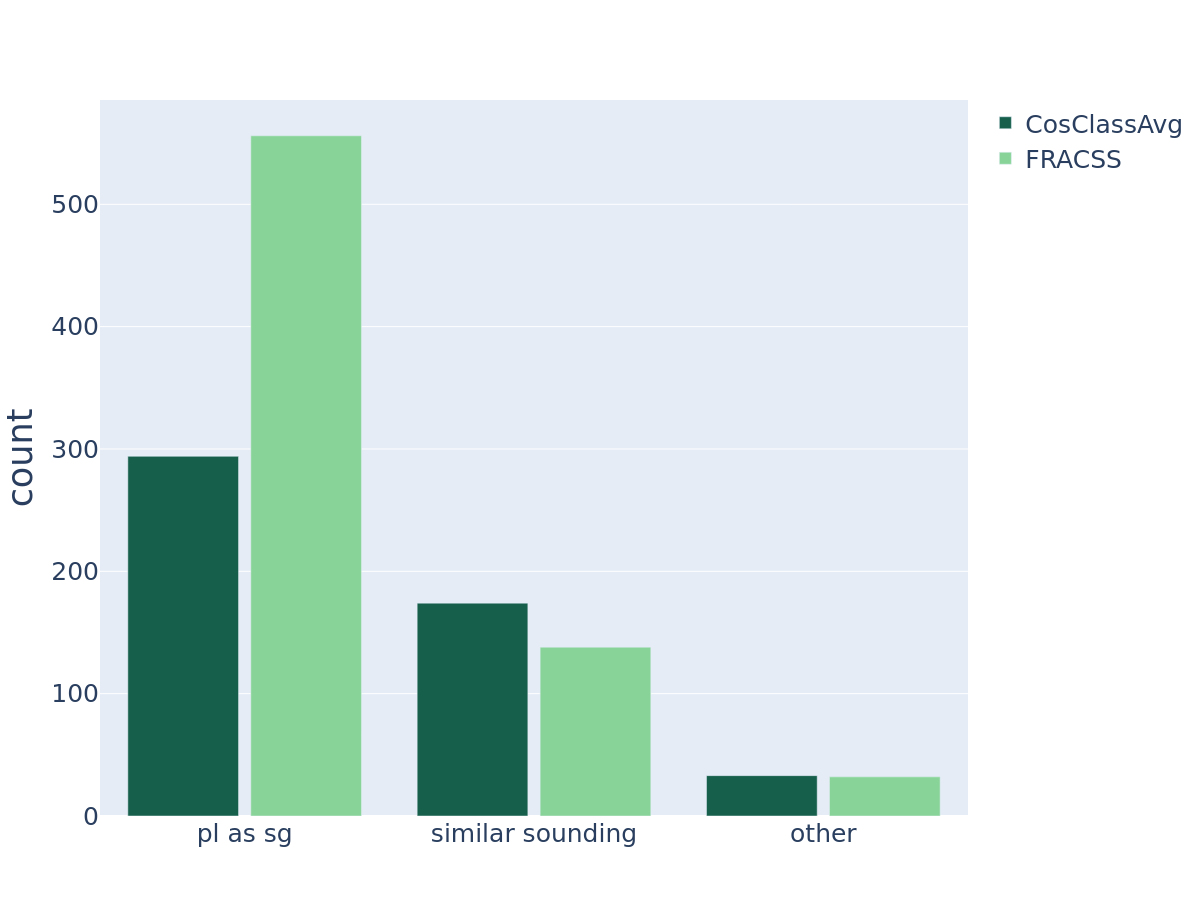}
    \caption{Different types of errors made by the DL-CosClassAvg ($N=501$) and the DL-FRACSS ($N=726$) model on the test set with 1034 tokens.}
    \label{fig:ldl-errors}
\end{figure}

We also examined the kind of errors made by the DL mappings for the words in the test data.  Overall, DL-FRACSS makes 726 errors in the evaluation of the test set, and DL-CosClassAvg  501 errors.  There are 439 word tokens that both models fail to predict correctly.  We distinguished between three types of errors, tabulated in Fig.~\ref{fig:ldl-errors}. First, many seen-stem plural words of the test set are recognized as their singular word. FRACSS tends to make more errors of this sort, for which both models frequently get the plural word as their second-best guess, and they always find the plural word among their first four guesses. Highly-ranked competitors tend to be synonyms or semantically related words.

Most of the remaining errors are observed for words with similar forms.  To assess this quantitatively, we computed the recall and the overlap indices between the set of target triphones $t$ and the set of predicted triphones $p$ as follows:
\begin{align*}
    \text{recall } (t, p) &= \frac{\lvert t \cap p\rvert}{\lvert p\rvert},\\
    \text{overlap } (t, p) &= \frac{\lvert t \cap p\rvert}{\min (\lvert t \rvert, \lvert p\rvert)}.\\
\end{align*}
For example, the word \textit{bribes} is recognized as \textit{tribes} by both models. The predicted and the target word share many form features with a recall and an overlap index of 0.6.  We classified words as `similar sounding' when the overlap index was greater than 0.3 and the recall index was greater than 0.2. The remaining words were assigned to the `other' class.   The set of words for which a similar-sounding error was made by DL-FRACSS is a subset of that of DL-CosClassAvg.  The two models are in error for the same 33 words assigned to the `other' category.

\subsection{Discussion} 

Both CosClassAvg and FRACSS generate high-quality plural vectors.  FRACSS plurals are somewhat better aligned with respect to angle, whereas CosClassAvg plurals are better positioned in terms of length.  We used the DL model to assess whether FRACSS or CosClassAvg vectors are closer aligned with words' forms.  On training data, both models have very similar performance.  On held-out test data, CosClassAvg is more accurate.  We take this as evidence that plural vectors generated by CosClassAvg are better aligned with the corresponding plural forms.

\section{General Discussion}\label{sec:general-discussion}

Using distributional semantics, visualization with t-SNE, and WordNet, we have documented for nearly 15,000 pairs of English singulars and their corresponding plurals that how plural semantics is realized in semantic space varies with the semantic class of the base word. Instead of there being one universal shift from singulars to plurals in distributional space, the direction and length of shift vectors depends on a lexeme's own semantics. As a consequence, shift vectors for fruits are substantially different from shift vectors for instruments.   

We proposed the CosClassAvg model to account for the conceptualization of a plural given the singular. This model proposes that an empirical plural vector is the sum of four vectors: the vector of the lexeme, the shift vector appropriate for its semantic class, a lexeme-specific vector representing the lexeme's own lexical properties, and an error vector representing measurement noise.  We showed that CosClassAvg provides more precise approximations of plural vectors than a model based on a general average shift vector (3CosAvg).  

We compared the CosClassAvg model with the FRACSS model \citep{Marelli:Baroni:2015}. The FRACSS model also takes the semantic vector of the singular as input, but makes use of matrix multiplication instead of vector addition to calculate the semantic vector of the plural.  The FRACSS model generates plural vectors that are closer to the target plural vectors. However, the plural vectors produced by FRACSS are shorter than the target plural vectors. 

To better understand the merits of the two models, we also considered how well the FRACSS plural vectors and the CosClassAvg vectors are aligned with words' form vectors.  We evaluated the quality of the alignment with the Discriminative Lexicon model \citep{Baayen:Chuang:Shafaei:Blevins:2019}, focusing on its comprehension network.  We created form vectors by first collecting all possible triphones and then specified, for a given word, in a high-dimensional binary vector, which triphones are present (1) in that word and which are absent (0).  We created two mappings from form vectors to semantic vectors, one for semantic vectors that use FRACSS to generate plural vectors, and a second mapping for semantic vectors that use CosClassAvg to produce plural vectors.  For training data, both types of vectors allowed highly accurate mappings to be set up.  However, for the held-out test data, plural vectors could be predicted with substantially higher accuracy when plural vectors were created using CosClassAvg.  This suggests that plural vectors created with CosClassAvg are better aligned with plurals' forms compared to vectors generated with FRACSS.  

CosClassAvg offers two advantages compared to FRACSS.  First, FRACSS models run the risk of overparameterization, especially for small datasets with numbers of observations that are substantially smaller than the square of the dimension of the semantic vectors. Second, the FRACSS matrix operation seems to suggest that pluralization is a unitary operation, represented by one transformation matrix.  However, what this model is actually doing is to capture, within one highly-parameterized mapping operation, a wide range of different ways in which plurals are realized, depending on the semantic class of their lexemes.   For comparison, one can set up a single FRACSS model for all bi-morphemic suffixed derived words of English, with high accuracy on both training and test data (see the supplementary materials for further details). However, the different derivational suffixes of English serve different semantic goals, which emerge immediately from a t-SNE visualization.  Thus, being able to obtain a high-quality mapping between singulars and plurals does not guarantee that the same semantic operation is governing all transitions from singulars to their plurals in semantic space.  

This conclusion has important consequences for the principle of semantic compositionality \citep{Pelletier2001} as applied to morphology.  According to this principle, the meaning of a plural is determined by the meaning of the singular and the meaning of the plural suffix, or the meaning that is realized by the rule that creates plurals from singulars. As we have seen, a general shift vector that is the same for all lexemes (as formalized by the 3CosAvg method) has some value, but fails to have the required precision.  FRACSS does not provide a uniform pluralization operation either, as, thanks to its large numbers of parameters, it can wrap itself around the many individual clusters of shift vectors that are characteristic of a large number of specific semantic classes.  It is, of course, possible to adjust the CosClassAvg model
$$
\bm{v}_{\text{pl}} = \bm{v}_{\text{sg}} + \bm{v}_{\textsc{pl} \mid \text{semantic class}} + \bm{\epsilon}_\text{lexeme} + \bm{\epsilon}
$$
by subtracting the average plural vector $\bar{\bm{v}}$ from all class-specific vectors:
$$
\bm{v}_{\text{pl}} = \bm{v}_{\text{sg}} + \bar{\bm{v}}  + [\bm{v}_{\textsc{pl} \mid \text{semantic class}} - \bar{\bm{v}}] + \bm{\epsilon}_\text{lexeme} + \bm{\epsilon}.
$$
This formulation of CosClassAvg isolates what is common to all plurals. Unfortunately, this common core is a shift vector that is located far outside the cluster of actual shift vectors (see Fig. \ref{fig:scatter-r-theta-shift}), and hence it remains unclear what is gained by incorporating it into the CosClassAvg model.  As a consequence, it is also unclear in what sense English plurals are `compositional' in the sense of, e.g., \cite{Pelletier:1994}.  At the same time, the present findings dovetail well with the insight from usage-based grammar and corpus linguistics that individual words, including inflected words, often have their own highly specific usage profiles \citep[see, e.g.,][]{Sinclair:1991}.

Noun pluralization has been characterized as being rather close to derivation: \cite{Booij:1996} characterizes it as inherent inflection, rather than contextual inflection.  It remains an issue for further research to clarify whether the present conclusions for nominal pluralization generalize to contextual inflection. Agreement marking on English simple present verbs makes for an interesting case to pursue in parallel with the present results on nominal plurals.

We conclude this study of the semantics of English noun pluralization by placing English in a broader cross-linguistic perspective.  Many languages have rules that are sensitive to semantic subsets of nouns.  Some languages split nouns into a group for which plurality marking is relevant, and a group for which it is irrelevant.  Typically, such splits are made along an animacy hierarchy, from kinship nouns at the highest rank, to human nouns, to (higher and lower) animate nouns, to inanimate nouns at the lowest rank \citep{Corbett:2000:Number}.  

In Slave, an Athabaskan language in Northwest Territories, Canada, plural marking occurs optionally only for human nouns and dogs \citep{Rice:1989:Slave}. The World Atlas of Language Structures documents 60 other languages that have an optional or obligatory plural marking for human nouns and lack a plural for nouns further down the animacy hierarchy \citep{WALS:34-Haspelmath:2013}.

In Persian, subject-verb agreement in person and number coded on the verb is obligatory for animate plural nouns but optional for inanimate ones \citep[][p.~145]{Mahootian:2002:Persian}. \cite{Smith-Stark:1974} reports a similar rule in Georgian.
Maori provides a case where number marking is obligatory only for kinship nouns such as \textit{matua} `parent' and \textit{teina} `younger sibling' \citep{Bauer:1993:Maori}.

Kiowa, an endangered Tanoan language spoken in Oklahoma, exhibits  a strong relationship between semantically coherent noun classes and number agreement behavior. Table~\ref{tab:kiowa-noun-classes} summarizes the nine classes distinguished by \cite{Harbour:2011, Harbour:2008:MorphosemanticNumber}, on the basis of which he argues for a \textit{morphosemantic theory} of number.  Bantu languages are known for their large numbers of semantically motivated noun classes \citep[see, e.g.,][for Swahili]{polome1967swahili}.   

\begin{table}[ht]
    \begin{center}
    \begin{minipage}{250pt}
    \caption{Kiowa noun classes based on \cite{Harbour:2011, Harbour:2008:MorphosemanticNumber}} \label{tab:kiowa-noun-classes}%
        \begin{tabular}{@{}clll@{}}
        \toprule
        Class & Semantic characteristics & Example \\
        \midrule
        1 & First person only  & `I' \\ 
        2 & Animates & `boy', `bird' \\ 
          & and independently mobile inanimates & `leg', `moon' \\ 
        3 & Default for vegetation & `grass' \\ 
          & and implements & `pencil' \\
        4 & Vegetation forming natural collections & `tree' \\ 
          & and implements that act collectively & `ember' \\ %
        5 & Hair types & `eyelash' \\ 
          & and midsize fruit growing in clusters & `tomato' \\
        6 & Individuable objects & `river' \\ 
        7 & Non-granular mass nouns & `water' \\ 
        8 & Pluralia tantum nouns, & `trousers' \\ 
          & composite nouns & `necklace' \\
          & and granular mass nouns & `rice' \\ 
        9 & Default & `shoe' \\ 
        \botrule
        \end{tabular}
    \end{minipage}
    \end{center}
\end{table}

English has in few instances grammaticalized the diverse ways in which our minds perceive and structure the objects and ideas in the world with which we interact. For English nouns, the distinction between mass and count nouns comes to mind. Additionally, a major part of present-day English count nouns that never or occasionally take the suffix \textit{-s} in their plural form are animal nouns that are hunted (e.g., \textit{duck}, \textit{woodcock}, and \textit{elk}) or fished (\textit{salmon} and \textit{crab}) \citep[see][for lexemes other than animal names]{Quirk:etal:85} \citep[see][for an extended list of 85 animal nouns]{Toupin:2015}. 

We kept the stimuli in the present study simple and consistent by focusing on regular singular and plural forms. 
Further research is required that investigates varieties of plurals including irregular plurals (e.g., \textit{man} $\sim$ \textit{men}), zero plurals (\textit{fish} $\sim$ \textit{fish}), pluralia tantum (\textit{scissors} with no singular variant), singularia tantum (\textit{wealth} with no plural variant), or sense-specific plural formations (\textit{mouse} $\sim$ \textit{mice} for rodents and \textit{mouse} $\sim$ \textit{mouses} for computer input devices; \citealp{Acquaviva:2008:LexicalPlurals}).

Many other languages reflect in their grammars a variety of ways in which nouns are perceived to cluster semantically.  Whereas semantic differentiation in the nominal system is explicitly grammaticalized in these languages, semantic noun classes also play a role in the grammar of English,  albeit mainly implicitly.  By combining distributional semantics, WordNet, and t-SNE visualization, we have been able to detect that semantic noun clusters also structure English language use.

\section*{Declarations}

\subsection*{Funding}

This research was funded by the European Research Council under the ERC grant number 742545, Project WIDE, awarded to the last author. The data processing of the NewsScape corpus was funded by the Competence Network for Scientific High Performance Computing in Bavaria to the third author.

\subsection*{Conflict of interest/Competing interests}
No potential conflict of interest was reported by the authors.




\subsection*{Authors' contributions}
E. S-B was responsible for data collection, analysis, interpretation, and wrote the first draft of the manuscript. M. M-T contributed to the~ data collection, analyses, and interpretation. P. U assembled the auditory resources, and contributed to data collection, interpretation, and writing. R. H. B contributed to the research planning, data analysis, interpretation, and writing.

\noindent

\bibliography{bibliography}


\end{document}